\documentclass[sigconf]{acmart}

\usepackage{algorithm}
\usepackage{algorithmic}

\usepackage{pifont}

\usepackage{multirow}
\usepackage{graphicx}
\usepackage{booktabs}
\usepackage{makecell}
\usepackage{xcolor}
\usepackage{vcell}
\usepackage{colortbl}
\usepackage{caption}
\usepackage{pifont}
\usepackage{adjustbox}
\usepackage[toc]{multitoc}
\usepackage[skins]{tcolorbox}
\usepackage[inline]{enumitem}
\usepackage{enumitem}
\usepackage{boldline}
\usepackage{hhline}
\usepackage{amsmath}
\usepackage{makecell}

\usepackage{newfloat}
\usepackage{listings}
\usepackage{utfsym}
\usepackage[normalem]{ulem}
\useunder{\uline}{\ul}{}
\usepackage[inline]{enumitem}
\usepackage{enumitem}
\usepackage{IEEEtrantools}
\usepackage{stmaryrd}

\definecolor{wkred}{RGB}{255, 190, 190}
\definecolor{wkblue}{RGB}{210, 230, 250}

\newcommand{\second}{\cellcolor{wkblue}}
\newcommand{\best}{\cellcolor{wkred}}

\AtBeginDocument{%
  }

\setcopyright{acmlicensed}
\copyrightyear{2025}
\acmYear{2025}
\acmDOI{XXXXXXX.XXXXXXX}

\acmConference[The Web Conference '2025]{Make sure to enter the correct
  conference title from your rights confirmation emai}{28 April - 2 May,
  2025}{Sydney, Australia}
\acmISBN{978-1-4503-XXXX-X/18/06}

\begin{document}

\title[A Chinese Multimodal Math Dataset To Evaluate and Enhance the Mathematics Reasoning of LMMs]{CMM-Math: A Chinese Multimodal Math Dataset To Evaluate and Enhance the Mathematics Reasoning of Large Multimodal Models}

\author{
    Wentao Liu$^\dag$, Qianjun Pan$^\dag$, Yi Zhang, Zhuo Liu, Ji Wu, Jie Zhou$^*$, Aimin Zhou$^*$, \\Qin Chen, Bo Jiang, Liang He \\
      $^1$School of Computer Science and Technology, East China Normal University, Shanghai, China\\ 
      $^2$Department of Educational Information Technology, East China Normal University, China \\ 
      $^3$Lab of Artificial Intelligence for Education, East China Normal University, Shanghai, China \\
    \url{https://github.com/ECNU-ICALK/EduChat-Math}
}

\renewcommand{\shortauthors}{Wentao Liu al.}

\begin{abstract}
Large language models (LLMs) have obtained promising results in mathematical reasoning, which is a foundational skill for human intelligence. Most previous studies focus on improving or measuring the performance of LLMs based on textual math datasets (e.g., MATH, GSM8K). 
Recently, a few researchers have released English multimodal math datasets (e.g., MATHVISTA and MATH-V) to evaluate the effectiveness of large multimodal models (LMMs). 
In this paper, we release a Chinese multimodal math (CMM-Math) dataset, including benchmark and training parts, to evaluate and enhance the mathematical reasoning of LMMs. 
CMM-Math contains over 28,000 high-quality samples, featuring a variety of problem types (e.g., choice, fill-in-the-blank, analysis) with detailed solutions across 12 grade levels from elementary to high school in China. 
Each problem may contain multiple images, and the visual context may be present in the questions or opinions, which makes this dataset more challenging.
Our comprehensive analysis reveals that state-of-the-art LMMs on the CMM-Math dataset face challenges, emphasizing the necessity for further improvements in LMM development. 
We also propose a Multimodal Mathematical LMM (\texttt{Math-LMM}) to handle the problems with mixed input of multiple images and text segments.
The \texttt{Math-LMM} is trained using three stages: foundational pre-training, foundational fine-tuning, and mathematical fine-tuning. 
The extensive experiments indicate that our model effectively improves math reasoning performance by comparing it with the SOTA LMMs over three multimodal mathematical datasets. 
Our datasets, codes, and weights will be released on GitHub.
\end{abstract}

\vspace{-3mm}


\vspace{-3mm}
\begin{teaserfigure}
    \vspace{-2mm}
    \hspace*{8mm} 
    \includegraphics[width=0.9\textwidth]{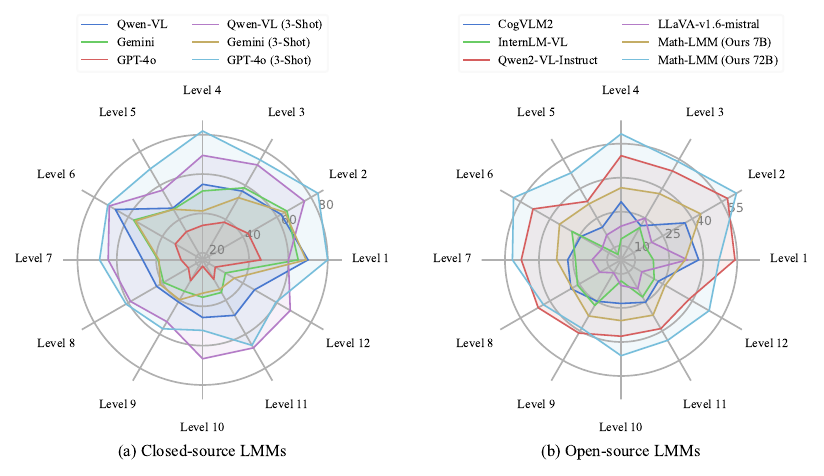}
    \vspace{-5mm}
    \caption{The performance of large multimodal models (LMMs) across 12 difficulty levels in our CMM-Math, where Level 1 is the easiest and Level 12 is the most difficult.}
    \Description{The performance of LMMs across 12 levels.}
    \label{fig:performance_level}
\end{teaserfigure}

\received{20 February 2007}
\received[revised]{12 March 2009}
\received[accepted]{5 June 2009}

\maketitle

\begin{figure*}[htbp]
\vspace{-2mm}
  \hspace*{1mm}
  \centering
  \includegraphics[width=0.95\textwidth]{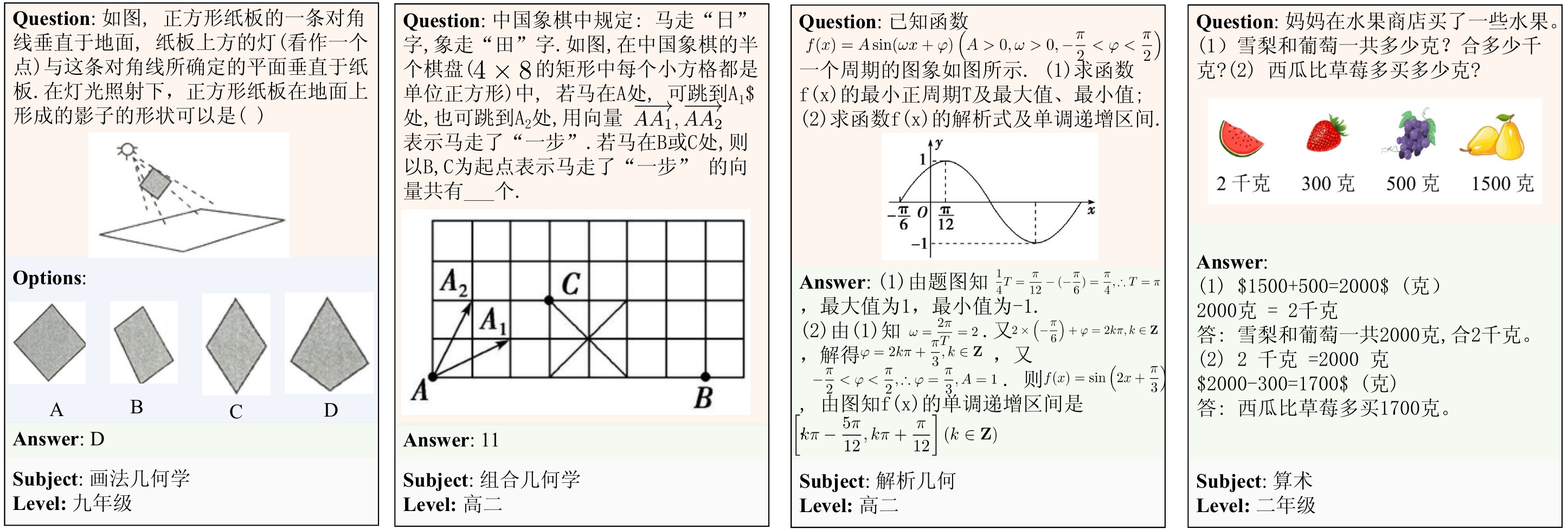}
  \vspace{-2mm}
  \caption{Some examples in our CMM-Math datasets.}
  \label{fig:example}
  \vspace{-3mm}
\end{figure*}

\section{Introduction}
\label{sec:Intro}
The recent advent of large language models (LLMs), like OpenAI's GPT-4 \cite{openai2023gpt4} and LLaMA \cite{touvron2023llama}, have demonstrated remarkable success across math problem solving, which is an important skill for human intelligence \cite{liu2023mathematical}.
Several studies utilize chain-of-thought (CoT) \cite{wei2022chain} to solve the problems step by step \cite{zhang2023automatic}. 
Also, tools and programs are integrated with LLMs to reduce the hallucination problem of math reasoning \cite{gao2023pal}. 
LLEMMA \cite{azerbayev8llemma} is a math-specific LLM that is fine-tuned on large-scale mathematical datasets, which obtains great abilities in generating textual solutions for mathematical problems and formulating formal proofs.

Furthermore, to evaluate and improve the mathematical performance of LLMs, some textual math datasets are released \cite{mishra2022lila,frieder2024mathematical}. 
For instance, \citet{hendrycks2measuring} proposed a MATH dataset that contains 12,500 challenging competition mathematics problems with step-by-step solutions.
GSM8K \cite{cobbe2021training} is a dataset of 8.5K high-quality linguistically diverse grade school math word problems.
However, these studies mainly focus on textual information while the visual contexts of the problems are not well studied by the previous literature.

Recently, various large multimodal models (LMMs) like GPT-4V \cite{openai2023gpt4}, LLaVA \cite{liu2024visual}, CogVLM \cite{wang2023cogvlm} and Gemini \cite{geminiteam2024geminifamilyhighlycapable} are proposed for improving the performance of multimodal reasoning. 
To measure the mathematic reasoning abilities of LMMs, some researchers released multimodal math benchmarks.
MMMU \cite{yue2024mmmu} is a popular multimodal dataset that mainly focuses on visual recognition with few problems about simple mathematical reasoning.
Additionally, the multimodal datasets designed for mathematical reasoning are also proposed \cite{lu2024mathvista,wang2024measuring}. 
\citet{lu2024mathvista} proposed the first math-specific English benchmark dataset, MATHVISTA, to measure the performance of the LMMs. 
This dataset includes seven mathematical reasoning types with diverse visual contexts across five primary tasks. 
Then, \citet{qiao2024we} collected 6.5K visual math problems to explore the problem-solving principles beyond the end-to-end performance. 
Additionally, \citet{wang2024measuring} organized a math evaluation benchmark, MATH-V, which contains 3,040 mathematical problems across 16 knowledge concepts.  
These studies mainly focus on measuring the performance of mathematical reasoning for the existing LMMs in English.

Unlike these studies, the unique characteristics of our Chinese Mathematical (CMM-Math) dataset are listed as follows.
First, the existing multimodal mathematical datasets are in English, while our CMM-Math focuses on Chinese. 
Second, we release a benchmark dataset to measure performance and contribute a training dataset to fine-tune LMMs, facilitating future research and increasing accuracy in multimodal mathematical reasoning.
Third, our dataset is more complex, with problems that may contain multiple figures in questions or options, as shown in the first example in Figure \ref{fig:example}. Moreover, it contains more question types, including choice problems, fill-in-the-blank problems, yes-no problems, and analysis problems.
Fourth, our large-scale CMM-Math datasets contain more than 28,000 samples across 12 grade levels from elementary to high school, to ensure the completeness and diversity of the dataset. 

We evaluate the state-of-the-art open-source and close-source LMMs on our CMM-Math dataset. 
The results show that it is a challenge for the current foundation LMMs to handle multimodal mathematical reasoning tasks.
Furthermore, we propose a math-specific LMM (\texttt{Math-LMM}) with a mixed instruction to handle the problems with multiple images. 
We train \texttt{Math-LMM} using three phases: foundational pre-training that aligns visual information with LLM, foundational fine-tuning that captures the abilities of task solving, and mathematical fine-tuning that learns mathematical reasoning.
We evaluate the performance of \texttt{Math-LMM} over three benchmark datasets and the results show our model outperforms the open-source LMMs in most cases.

The main contributions of this paper are summarized as follows.
\begin{itemize}[leftmargin=*, align=left]
    \item We collect and release a high-quality Chinese multimodal mathematical dataset, CMM-Math, which consists of evaluation and training datasets to measure and improve the performance of the current LMMs.
    \item We propose a math-specific LMM (\texttt{Math-LMM}) that trains with three stages: foundational pre-training, foundational fine-tuning, and mathematical fine-tuning. 
    \item We conduct a comprehensive evaluation of the existing LMMs and our \texttt{Math-LMM} model on the CMM-Math dataset. The results show that multimodal mathematical reasoning is still a great challenge for LLMs. Our model outperforms the strong open-source LMMs over CMM-Math, MATHVISTA and Math-V datasets in most cases. 
\end{itemize}


\vspace{-3mm}
\section{Related Works}
\label{sec:Related}
\vspace{-1mm}
\subsection{Mathematical Dataset}
\vspace{-1mm}
Various datasets \cite{mishra2022lila,cobbe2021training,hendrycks2021measuring,frieder2024mathematical} have been introduced to measure the mathematical capabilities of large language models (LLMs). 
For instance, GSM8K \cite{cobbe2021training} and MATH \cite{hendrycks2021measuring} are two popular textual datasets to comprehensively evaluate the LLMs.
These datasets \cite{liu2023mathematical} can test the language processing abilities in areas such as mathematical reasoning, computation, and theorem proving, but they do not effectively assess the ability of large multimodal models to solve math problems that require visual modalities, such as geometry problems, properties of function graphs, etc.

In the past two years, specialized multimodal datasets for the mathematics domain have emerged to assess the mathematical capabilities of large multimodal models, such as MMMU \cite{yue2024mmmu}, MATHVISTA \cite{lu2024mathvista}, We-Math \cite{qiao2024we} and MATH-V \cite{wang2024measuring}. MMMU \cite{yue2024mmmu} primarily evaluates the model's visual recognition abilities, with only a few questions involving simple mathematical reasoning. MATHVISTA \cite{lu2024mathvista} consolidates and transforms existing FQA (Figure Question Answering), GPS (Geometry Problem Solving), MWP (Math Word Problems), TQA (Textbook Question Answering), and VQA(Visual Question Answering) datasets, enabling tests of basic mathematical reasoning, but it covers fewer math concepts and has a limited variety of question types, many of which can be solved with VQA capabilities alone. MATH-V \cite{wang2024measuring} compiles questions from math competitions like Math Kangaroo, UK contests at various levels (Grey, Pink, Junior, Senior), and US competitions (AMC 8, 10, 12) and invitational events (AIME). This dataset evaluates models across multiple educational levels and topics. 

Unlike these datasets, our CMM-Math dataset not only offers evaluations covering multiple grades and topics but also includes both training and evaluation datasets. Furthermore, the problem in our dataset may contain multiple images with detailed solution explanations across a variety of problem types. We have also provided a large amount of Chinese multimodal problems with multiple images across 12 grade levels from elementary to high school.                                     

\vspace{-2mm}
\subsection{Large Multimodal Model}
\vspace{-1mm}
With the introduction of GPT-4V \cite{openai2023gpt4}, many versatile large multimodal models have been proposed. Commonly, these models aim to handle both text and image tasks simultaneously. They employ a tokenizer and a visual encoder (such as CLIP \cite{radford2021learning}) to encode textual and visual information separately, then concatenate the encoded vectors into a unified input that is fed into the large language model. Notable LLMs include LLaVa \cite{liu2024visual}, Qwen-VL-Max \cite{bai2023qwenvlversatilevisionlanguagemodel}, Gemini \cite{geminiteam2024geminifamilyhighlycapable}, InternLMX Composer-VL \cite{internlmxcomposer2_5}, and GPT-4o. 
These models have greatly succeeded in general multimodal tasks, such as OCR, visual question answering, image captioning, multimodal reasoning, etc. Our CMM-Math aims to provide a detailed and comprehensive evaluation of the mathematical multimodal reasoning capabilities of these models.
Unlike training methods for general-purpose large multimodal models, we propose a three-stage training method specifically for the mathematical domain. 

\begin{table}[t!]
\centering
\caption{Key statistics of CMM-Math. }
\label{table: Overview}
\vspace{-3mm}
\begin{tabular}{ll}
\toprule
\textbf{Statistic} & \textbf{Number} \\
\midrule

Total problems  &   28,069   \\
Total images &  15,213  \\
Total detailed solutions &  23,825  \\
\midrule
Levels &  12  \\ 
Subjects &  13 \\
Images in questions &  9,490   \\
Images in answers &  5,723   \\
Maximum problem length & 2,016 \\
Minimum problem length & 3 \\
Average problem length & 108.31 \\
\bottomrule
\end{tabular}
\vspace{-3mm}
\end{table}

\begin{table*}[t!]
\centering
\small
\caption{Comparison with existing datasets. EN and CN mean English and Chinese.}
\label{table: comparison}
\vspace{-3mm}
\setlength{\tabcolsep}{1.6mm}{
\begin{tabular}{lllllll}
\hline
          & Multimodal & Training & Evaluation & Language & \#Number & Type \\ \hline
MATH \cite{hendrycks2021measuring}      &       \ding{55}          &    \ding{51}      &     \ding{51}       &      EN    &    12,500      &   Fill-in-the-blank questions  \\
GSM8K \cite{cobbe2021training}    &      \ding{55}      &    \ding{55}      &     \ding{51}       &     EN     &     8,500     &   Fill-in-the-blank questions   \\
MathQA \cite{amini2019mathqa}   &     \ding{55}         &    \ding{51}  &      \ding{51}       &      EN    &      37,259     &   Fill-in-the-blank questions   \\
MMMU \cite{yue2024mmmu}     &     \ding{51}         &    \ding{55}        &      \ding{51}        &      EN    &         11,500 &    Choice and fill-in-the-blank questions  \\
MATHVISTA \cite{lu2024mathvista} &      \ding{51}        &     \ding{55}      &       \ding{51}      &     EN     &      6,141    &   Choice and fill-in-the-blank questions   \\
We-Math \cite{qiao2024we} &      \ding{51}        &     \ding{55}      &       \ding{51}      &     EN     &      6,500    &   Choice questions   \\
MATH-V \cite{wang2024measuring}   &     \ding{51}         &     \ding{55}      &      \ding{51}       &     EN     &   3,040       &   Choice and fill-in-the-blank questions   \\
\hline
CMM-Math (Ours)  &     \ding{51}       &      \ding{51}      &      \ding{51}        &   CN       &    28,249     &  \makecell*[l]{Choice, fill-in-the-blank, \\yes-no and analysis questions}  \\
\hline
\end{tabular}}
\vspace{-2mm}
\end{table*}

\vspace{-3mm}
\section{Dataset}
\label{sec:Dataset}
\vspace{-1mm}
\subsection{Dataset Construction}
\vspace{-1mm}
Due to the lack of Chinese multimodal mathematical datasets, we collect and release the CMM-Math dataset, constructed with three main steps: data collection, data cleaning, and data annotation.

During the data collection phase, we gathered over 10,000 real exam papers covering 12 grades of mathematics test problems, from primary school to high school in China. Each exam paper includes various types of questions such as multiple-choice, fill-in-the-blank, and analysis problems. These questions contain both visual and textual information, along with their answers, solutions, grades, and question types. Since the exam papers are in PDF format, we used the Mathpix API\footnote{\url{https://mathpix.com}} to extract the text and images into markdown format and downloaded the extracted images locally. 

During the data cleaning phase, we first convert the problems from the markdown text into a JSON format, including fields such as question type, modality, images, question, options, answer, and solution. Then, we check and correct issues related to text and image recognition, such as text or mathematical formulas that are incorrectly identified as images, and images that are misrecognized. Finally, we detect and resolve potential issues within each field, including mismatched question types and overly long parsing.

During the data annotation phase, we ask students to check the problems with the format and context to ensure their quality. Then, we divide the problem into 13 different subjects. Due to the large amount of data and the high cost of manual annotation, we chose three large models, GPT-4o, Gemini, and Qwen-VL-max, to vote on the topics and assign them to the topics based on the principle of minority obeying majority.

\begin{table}[t!]
\small
\centering
\caption{Detailed statistics of CMM-Math datasets. }
\label{table: detailed statics}
\vspace{-2mm}
\setlength{\tabcolsep}{0.8mm}{
\begin{tabular}{llll}
\toprule
\textbf{Statistic} & \textbf{\#Evaluation}& \textbf{\#Training} & \textbf{\#Total} \\
\midrule
Total problems  & 5,821 & 22,248 & 28,069  \\
Total images & 3,794    & 11,419 & 15,213   \\
Total detailed solutions & 5,204    & 18,621 & 23,825   \\
\midrule
Images in questions & 2,144(56.51\%) & 7,346(64.33\%) & 9,490(62.38\%)\\
Images in answers & 1,650(43.49\%) & 4,073(35.67\%) & 5,723(37.62\%)\\
\midrule
Type    &    4    &    4    &    4\\
- Choice  &  2,222(38.17\%) & 8,618(38.74\%) & 10,840 (38.62\%)\\
- fill-in-the-blank  & 1,668(28.65\%) & 6,382(28.69\%)& 8,050(28.68\%)\\
- Yes-no  & 18(0.31\%)  & 88 (0.40\%) & 106(0.38\%)\\
- Analysis  & 1,913(32.86\%) & 7,170(32.23\%) & 9,083 (32.36\%)\\
\midrule
Level    &  12  & 12 &    12\\
- Level-1   & 319(5.48\%) & 1,180(5.30\%) & 1,499(5.34\%)\\
- Level-2   & 439(7.54\%) & 1,648(7.41\%)& 20,87(7.44\%)\\
- Level-3   & 444(7.63\%) & 1,680(7.55\%) & 2,124(7.57\%)\\
- Level-4   & 574(9.86\%) & 2,210(9.93\%) & 2,784(9.92\%)\\
- Level-5   & 534(9.17\%) & 1,939(8.72\%) & 2,473(8.81\%)\\
- Level-6   & 463(7.95\%) & 1,783(8.01\%) & 2,246(8.00\%)\\
- Level-7   & 458(7.87\%) & 1,751(7.87\%) & 2,209(7.87\%)\\
- Level-8   & 361(6.20\%) & 1,372(6.17\%)& 1,733(6.17\%)\\
- Level-9   &  493(8.47\%) & 1,900(8.54\%) & 2,393(8.53\%)\\
- Level-10   & 587(10.08\%) & 2,284(10.27\%) & 2,871(10.23\%)\\
- Level-11   & 646(11.10\%) & 2,512(11.29\%) & 3,158(11.25\%)\\
- Level-12   & 503(8.64\%) & 1,989(8.94\%) & 2,492(8.88\%)\\
\midrule
Subjects    &  13  &    13 &    13   \\
- Analytic Geometry   & 707(12.15\%) & 2,756(12.39\%) & 3,463(12.34\%)\\
- Metric Geometry   & 738(12.68\%) & 2,876(12.93\%) & 3,614(12.88\%)\\
- Solid Geometry   &  546(9.38\%) & 2,092(9.40\%) & 2,638(9.40\%)\\
- Arithmetic   & 1,999(34.34\%) & 7,855(35.31\%) & 9,854(35.11\%)\\
- Algebra   & 676(11.61\%) & 2,640(11.87\%) &  3,316(11.81\%)\\
- Counting   & 407(6.99\%) & 1,546(6.95\%) & 1,953(6.96\%)\\
- Transformation Geometry   & 85(1.46\%) & 274(1.23\%) & 359(1.28\%)\\
- Graph Theory   & 26(0.45\%) & 44(0.20\%) & 70(0.25\%)\\
- Combinatorial Geometry   & 140(2.41\%) & 495(2.22\%) & 635(2.26\%)\\
- Combinatorics   & 217(3.73\%)  & 747(3.36\%) & 964(3.43\%)\\
- Logic   & 127(2.18\%) & 416(1.87\%) & 551(1.96\%)\\
- Descriptive Geometry   & 135(2.32\%) & 465(2.09\%) & 603(2.15\%)\\
- Statistics   & 18(0.31\%) & 42(0.19\%) & 60(0.21\%)\\
\bottomrule
\end{tabular}}
\vspace{-4mm}
\end{table}

\subsection{Dataset Analysis}
Our CMM-Math dataset contains 28,069 problems with rich textual and visual information (See Table \ref{table: Overview} and Table \ref{table: detailed statics}). 
It contains 21,200 textual problems and 6,869 multimodal problems, divided into choice, fill-in-the-blank, yes-no, and analysis problems.
We also split the data into 12 levels, corresponding to the basic education stage from the first grade of primary school to the third grade of high school, to ensure the dataset's applicability and reference value in educational practice.
CMM-Math contains 13 knowledge points, covering most of the mathematical fields encountered in middle and high school, particularly logic, algebra, counting, arithmetic, combinatorics, graph theory, topology, statistics, solid geometry, metric geometry, analytic geometry, descriptive geometry, combinatorial geometry, and transformation geometry.

Our dataset has a more diverse and comprehensive range of question types compared to existing datasets. 
Particularly, the dataset is finely classified based on three dimensions: grade, subject, and question type. 
More than 84\% of problems have detailed solutions. 
Following the 1:4 ratio principle, we split the dataset into evaluation and training datasets to measure and enhance the performance of LMMs. 
The evaluation dataset contains 5,821 examples, and the training dataset contains 22,248 examples.

\vspace{-2mm}
\subsection{Comparison with Existing Datasets}
To clarify the characteristics of our CMM-Math, we compare it with multiple existing mathematical benchmarks, including textual datasets (e.g., MATH, GSM8K, MathQA) and multimodal datasets (e.g., MMMU, MATHVISTA, MATH-V, and We-Math), as shown in Table \ref{table: comparison}. 
First, most of these datasets are in English while we focus on Chinese to measure LMMs more comprehensively. 
Second, our CMM-Math is a multimodal dataset for mathematical reasoning, where each problem may contain multiple images, designed for LMMs. Textual MATH and GSM8K datasets are used to evaluate the performance of LLMs. 
Third, our dataset contains the evaluation and training datasets. The existing multimodal datasets evaluate the power of LMMs while ignoring how to improve their performance.
Fourth, our datasets are diverse, with rich subjects and problem types across 12 grades, which can more deeply distinguish models' performance in different mathematical contexts.
CMM-Math provides 11 knowledge points and multiple problem types, including choice, yes-no, fill-in-the-blank, and analysis. 
In addition, the CMM-Math problems are derived from real math tests in primary, middle, and high school, covering a wider range of grades than the GSM8K, which only includes primary school knowledge. CMM-Math can better test the problem-solving ability of the model at different stages. 
Finally, our dataset has detailed analysis and rich question stem content, which can enhance the mathematical reasoning ability of the model.

\begin{figure*}[htbp]
\vspace{-1mm}
  \centering
  \includegraphics[width=1.0\textwidth]{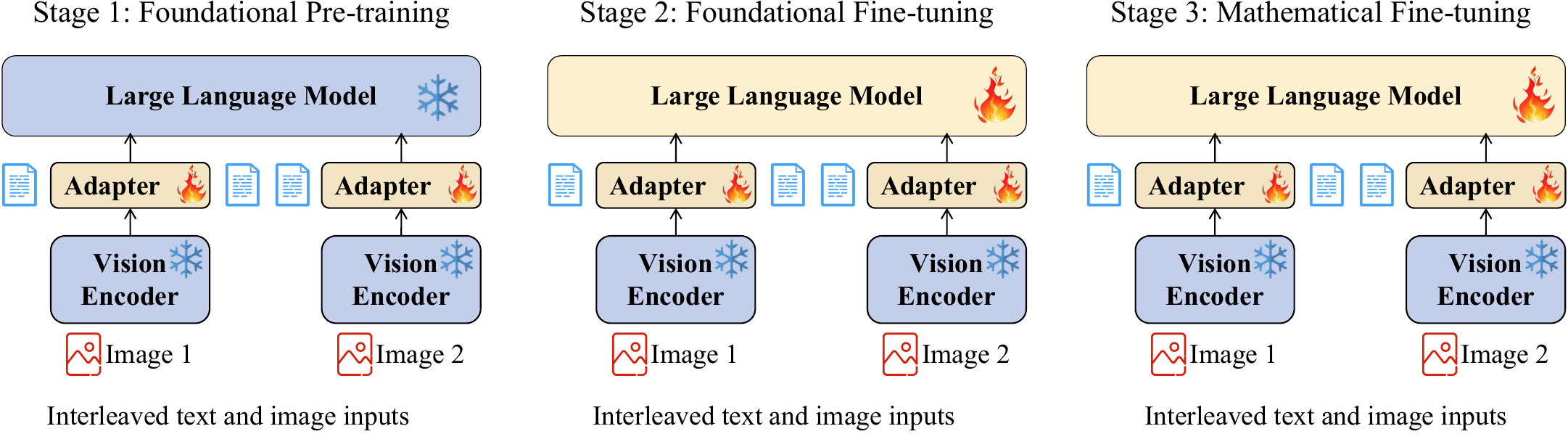}
  \vspace{-7mm}
  \caption{The diagram of our \texttt{Math-LMM} framework.}
  \label{fig:model_framework}
  \vspace{-3mm}
\end{figure*}

\vspace{-2mm}
\section{Our Proposed Method}
In this paper, we propose a math-specific LMM (\texttt{Math-LMM}) as a strong baseline for multimodal math reasoning, as illustrated in Figure \ref{fig:model_framework}. 
To tackle the problems with multiple figures, we introduce a mixed instruction with the interleaved text and image inputs. 
Inspired by LLaVa \cite{liu2024visual}, \texttt{Math-LMM} primarily comprises a Vision Encoder for encoding image information, an Adapter for modality alignment, and a Large Language Model (LLM) for mathematical reasoning. 
Moreover, to capture the abilities of math reasoning, we train our model using three stages: 1) the foundation pre-training stage aligns general visual information with LLMs via text-image pairs; 2) the foundation fine-tuning stage learns general multimodal abilities based on foundation instructions; and 3) the mathematical fine-tuning stage uses instruction learning based on math instructions to learn multimodal mathematical reasoning.

We translate the input sample as a mix instruction to train \texttt{Math-LMM}. The template instruction is ``This is text. [IMAGE1]. This is text. ... This is text. [IMAGE2]. [IMAGE3] ... ", where [IMAGE] is the image embedding after alignment.
First, we input the images into the vision encoder (e.g., ViT \cite{DBLP:conf/iclr/DosovitskiyB0WZ21}, DFN-5B-H-14+ \cite{fang2023data}) to obtain the image representation with a fixed dimension. The vision encoder is pre-trained on large-scale visual or multimodal datasets, which can embed the image effectively.
Then, we obtain the image embedding [IMAGE] by aligning the image representation with LLM using an adapter. Common designs for the adapter include Cross-Attention \cite{gheini2021cross}, Q-Former \cite{li2023blip}, and MLP. We chose a simple but equally effective two-layer MLP with GELU functions for our Adapter. 
Finally, we input the mixed restructure with interleaved image-text inputs to LLMs (e.g., Qwen2-7B-Instruct \cite{qwen}).


In the first foundational pre-training phase, we pre-train the adapter to align the image input with the LLM using large-scale general multimodal datasets with image descriptions. 
In particular, we only update the parameters of the adapter and fix the parameters of the LLM and encoder.
Here, we select several datasets for adapter pre-training, including LLaVA-Pretrain and LLaVA-CC3M-Pretrain-595K \cite{liu2024visual}, and Mantis-Instruct \cite{Jiang2024MANTISIM}. 

In the foundational fine-tuning phase, we focus on learning task processing capability using instruction tuning. We train the parameters of both the Adapter and the LLM modules via large-scale foundation instructions. 
We primarily use datasets involving general domain problems, such as ShareGPT-4o\footnote{\tiny \url{https://huggingface.co/datasets/OpenGVLab/ShareGPT-4o}}, MMDU \cite{liu2024mmdumultiturnmultiimagedialog}, llava-en-zh-300k \cite{liu2024visual}, and CogVLM-SFT-311K \cite{zhu2023minigpt,liu2024visual}.

In the third mathematical fine-tuning phase, we fine-tune the proposed \texttt{Math-LMM} for improved mathematical capabilities by adjusting both the Adapter and LLM modules. 
We train our model on mathematics-related datasets, including our proposed CMM-Math, GSM8K \cite{cobbe2021training}, competition\_math \cite{hendrycksmath2021}, blossom-math-v4\footnote{\tiny \url{https://huggingface.co/datasets/Azure99/blossom-math-v4}}, Vietnamese-395k-meta-math-MetaMathQA-gg-translated
, Vietnamese-meta-math-MetaMathQA-40K-gg-translated
, and Vietnamese-microsoft-orca-math-word-problems-200k-gg-translated
\footnote{\tiny \url{https://huggingface.co/datasets/5CD-AI}}
.


\begin{figure}[t!]
  \centering
  \includegraphics[width=0.45\textwidth]{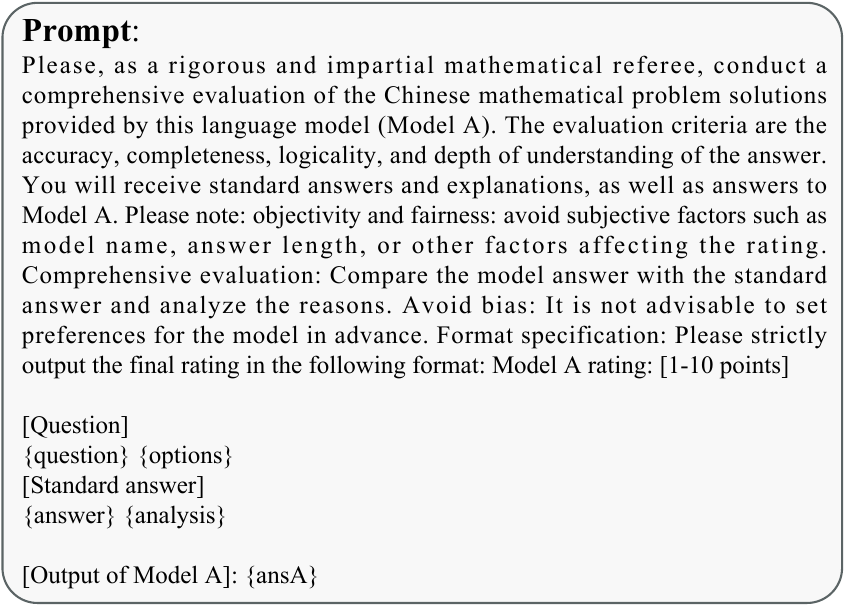}
  \vspace{-3mm}
  \caption{The prompt for GPT-4o used for scoring.}
  \vspace{-4mm}
\end{figure}

\begin{table*}[t!]
\centering
\small
\caption{Comparison of model performances in \textbf{accuracy} across different levels. The levels from 1 to 12 correspond to grades from primary to high school. 
The \colorbox{wkred}{first} and \colorbox{wkblue}{second} highest accuracy of open-source LMMs are marked in red and blue.
}
\label{table: comparison_accuracy_level}
\vspace{-2mm}
\setlength{\tabcolsep}{2.3mm}{
\begin{tabular}{l|c|cccc cccc cccc}
\toprule
Models & Overall & LV1 & LV2 & LV3 & LV4 & LV5 & LV6 & LV7 & LV8 & LV9 & LV10 & LV11 & LV12  \\
\midrule
\multicolumn{14}{c}{Open-source LMMs}\\
\midrule
CogVLM2   & 25.85  & 38.00  & 36.36  & 21.15  & 29.38  & 20.39  & 24.32  & 27.32  & 29.33  & 24.90  & 23.02  & 25.37  & 22.73 \\
InternLM-VL  & 19.82  & 18.00  & 15.91  & 20.19  & 12.99  & 6.58  & 28.83  & 23.90  & 26.00  & 26.97  & 13.06  & 22.69  & 20.63  \\
Qwen2-VL-Instruct  & 43.04 & \best{54.00} & \second{57.95}& \second{49.04}& 49.72& 33.55& 48.65& \second{47.80}& \best{46.00}& \second{41.08}& 37.46& \second{38.81}& 38.46 \\ 
LLaVA-v1.5 & 18.08 & 40.00& 27.27& 27.88& 21.47& 17.76& 25.23& 15.12& 18.67& 18.67& 15.81& 12.84& 9.09\\
LLaVA-v1.6-mistral  & 16.83 & 33.00& 19.32& 25.00& 18.64& 16.45& 13.51& 16.59& 14.67& 10.37& 15.12& 18.51& 14.34\\
CogVLM2 (3-Shot)   & 31.21  & 37.00  & 34.09  & 40.38  & 29.94  & 28.95  & 35.14  & 34.63  & 32.67  & 28.22  & 28.52  & 32.84  & 25.52 \\
InternLM-VL (3-Shot) & 25.09  & 22.00  & 32.95  & 27.88  & 16.95  & 11.84  & 27.93  & 25.37  & 26.67  & 24.90  & 24.05  & 26.87  & 31.82  \\
Qwen2-VL-Instruct (3-Shot)  & \second{46.29} & 38.00& \best{62.50}& \best{53.85}& \second{55.93}& \best{54.61}& \best{60.36}& \second{47.80}& \best{46.00}& \best{44.81}& \second{40.89} & 36.42& \second{43.01} \\
LLaVA-v1.5 (3-Shot) & 19.69 & 37.00& 28.41& 37.50& 24.86& 24.34& 30.63& 11.22& 16.67& 9.54& 19.93& 13.13& 18.18\\
LLaVA-v1.6-mistral (3-Shot)  & 21.88 & 17.00& 35.23& 31.73& 32.20& 26.97& 30.63& 20.00& 17.33& 7.47& 16.84& 24.18& 21.68\\

\midrule
\texttt{\textbf{Math-LMM}} (Ours 7B)  & 32.10 & 32.00 & 44.32 & 37.50 & 35.59 & 32.24 & 35.14 & 32.20 & 28.67 & 32.37 & 30.58 & 31.94 & 26.22\\
\texttt{\textbf{Math-LMM}} (Ours 72B)  & \best{48.57} & \second{47.00} & \best{62.50} & \best{53.85} & \best{59.32} & \second{48.03} & \second{58.56} & \best{51.71} & \second{43.33} & 38.59 & \best{46.05} & \best{44.78} & \best{48.60}\\
\midrule
\multicolumn{14}{c}{Closed-source LMMs}\\
\midrule
Qwen-VL-Max & 49.91  & 70.00  & 62.50  & 56.73  & 54.80  & 46.71  & 67.57  & 46.83  & 43.33  & 41.08  & 45.70  & 48.96  & 46.85  \\
Gemini & 41.88  & 65.00  & 65.91  & 58.65  & 51.41  & 46.05  & 56.76  & 39.02  & 39.33  & 34.85  & 35.40  & 35.52  & 29.72  \\
GPT-4o  & 29.02  & 46.00  & 43.18  & 38.46  & 33.90  & 32.89  & 32.43  & 27.32  & 24.67  & 28.63  & 19.59  & 27.76  & 23.78  \\
Qwen-VL-Max (3-Shot) & 64.91  & 60.00  & 76.14  & 72.12  & 69.49  & 57.24  & 71.17  & 64.39  & 58.67  & 52.70  & 66.67  & 68.06  & 67.83\\
Gemini (3-Shot) & 41.65  & 69.00  & 64.77  & 52.88  & 41.24  & 46.05  & 55.86  & 38.54  & 41.33  & 39.83  & 33.33  & 33.73  & 34.97  \\
GPT-4o (3-Shot) & 65.98 & 80.00 & 84.09  & 75.00  & 81.92  & 69.74  & 72.07  & 68.78  & 61.33  & 56.85  & 52.23  & 66.57  & 59.44  \\
\midrule
Mean accuracy of LMMs &35.66 & 44.61 & 47.41 & 43.32 & 39.99 & 34.47 & 43.04 & 35.47 & 34.15 & 31.21 & 31.35 & 33.83 & 32.38 \\
\bottomrule
\end{tabular}
}
\vspace{-2mm}
\end{table*}

\begin{table*}[t!]
\centering
\scriptsize
\caption{Comparison of model performances in \textbf{accuracy} across various mathematical subjects. Alg: algebra, AnaG: analytic geometry, Ari: arithmetic, CombG: combinatorial geometry, Comb: combinatorics, Cnt: counting, Desc: deseriptive geometry, GrphT: graph theory, Log: logic, MetG: metric geometry, SolG: solid geometry, Stat: statistics, TransG: transformation geometry. 
}
\label{table: comparison_accuracy_subject}
\vspace{-2mm}
\setlength{\tabcolsep}{1.4mm}{
\resizebox{\linewidth}{!}{
\begin{tabular}{l|c|cccc ccccc ccccc}
\toprule
Models  & Overall & Alg & AnaG & Ari & CombG & Comb & Cnt & Desc & GrphT & Log & MetG & SolG & Stat & TransG \\
\midrule
\multicolumn{15}{c}{Open-source LMMs}\\
\midrule
CogVLM2 & 25.85  & 21.94  & 25.22   &24.76  & 28.85  & 24.68  & 26.54  & \second{29.79}  & \best{44.44}  & 19.05  & 33.79  & 23.36  & 45.45  & 25.58 \\
InternLM-VL  & 19.82  & 21.94  & 26.38  & 20.68  & 21.15  & 12.99  & 16.67  & 19.15  & 0.00  & 9.52  & 20.48  & 12.15  & 27.27  & 13.95 \\
Qwen2-VL-Instruct & 43.04 & 36.45& \second{42.03}& 50.65& \second{36.54}& \second{40.26}& 45.68& 27.66& \best{44.44}& 39.68& 43.34& \second{37.85}& 45.45& \second{37.21} \\
LLaVA-v1.5 & 18.08 & 15.16& 13.91& 21.82& 15.38& 18.18& 12.96& 21.28& 22.22& 12.70& 19.45& 19.16& 36.36& 25.58\\
LLaVA-v1.6-mistral & 16.83 & 17.10& 16.52& 22.31& 7.69& 11.69& 10.49& 8.51& 0.00& 19.05& 13.99& 15.42& 27.27& 16.28\\
CogVLM2 (3-Shot) &31.21  & 30.97  &28.12  & 32.57  &32.69  & 24.68  & 30.25 & \second{29.79}  &22.22   &  25.40  & 36.18  & 30.84 & 45.45  & 27.91 \\
InternLM-VL (3-Shot)& 25.09  & 29.03  & 30.14  & 24.76  & 15.38  & 27.27  & 25.93  & 12.77  & 0.00  & 26.98  & 24.57  & 17.76  & 9.09  & 25.58 \\
Qwen2-VL-Instruct (3-Shot)  & \second{46.29} & \second{37.42}& 40.58& \second{53.42}& \best{46.15}& \second{40.26}& \best{56.17}& \best{36.17}& \second{33.33}& \second{46.03}& \best{49.15}& \best{40.19}& \second{54.55}& \best{51.16} \\
LLaVA-v1.5 (3-Shot)  & 19.69 & 19.03& 14.20& 28.66& 3.85& 23.38& 21.60& 14.89& 11.11& 15.87& 14.68& 16.36& 45.45& 2.33\\
LLaVA-v1.6-mistral (3-Shot)  & 21.88 & 22.26& 17.97& 32.25& 5.77& 20.78& 22.22& 4.26& 11.11& 28.57& 15.70& 13.55& 27.27& 16.28\\
\midrule

\texttt{\textbf{Math-LMM}} (Ours 7B) & 32.10 & 26.13 & 29.86 & 36.64 & 28.85 & 32.47 & 35.19 & 27.66 & \second{33.33} & 30.16 & 35.15 & 28.04 & 36.36 & 25.58\\
\texttt{\textbf{Math-LMM}} (Ours 72B)  & \best{48.57} & \best{48.06}& \best{44.35}& \best{60.59}& 32.69& \best{50.65}& \second{53.70}& 17.02& 22.22& \best{57.14}& \second{44.03}& 35.98& \best{63.64}& 27.91\\

\midrule
\multicolumn{15}{c}{Closed-source LMMs}\\
\midrule
Qwen-VL-Max  & 49.91  & 49.68 & 48.12  & 60.59  & 44.23  & 50.65  & 54.94  & 21.28  & 11.11  & 50.79  & 42.66  & 36.45  & 63.64  & 51.16 \\
Gemini & 41.88  & 36.77  & 37.97  & 55.37  & 26.92  & 37.66  & 48.15  & 31.91  & 33.33  & 26.98  & 37.54  & 32.71  & 54.55  & 25.58 \\
GPT-4o  & 29.02  & 26.13  & 28.41  & 35.67  & 26.92  & 20.78  & 32.10  & 25.53  & 33.33  & 28.57  & 24.57  & 21.96  & 18.18  & 37.21 \\
Qwen-VL-Max (3-Shot) & 64.91  & 70.00  & 63.19  & 74.43  & 50.00  & 66.23 & 69.75  & 21.28  & 44.44  & 73.02  & 54.61  & 57.01  & 63.64  & 53.49 \\
Gemini (3-Shot) & 41.65  & 38.71  & 35.07  & 54.07  & 19.23  & 31.17  & 51.85  & 27.66  & 44.44  & 44.44  & 36.18  & 30.84  & 54.55  & 44.19 \\
GPT-4o (3-Shot)  & 65.98  & 62.90  & 59.71  & 82.57  & 46.15  & 66.23  & 72.84  & 40.43  & 77.78  & 68.25  & 59.04  & 47.66  & 81.82  & 55.81 \\
\midrule
Mean accuracy of LMMs &35.66 & 33.87 & 33.43 & 42.88 & 27.14 & 33.33 & 38.17 & 23.17 & 27.16 & 34.57 & 33.62 & 28.74 & 44.44 & 31.27 \\
\bottomrule
\end{tabular}}
}
\vspace{-2mm}
\end{table*}

\begin{table*}[t!]
\centering
\small
\caption{Comparison of model performances in \textbf{GPT-4o score} across different levels. The levels from 1 to 12 correspond to primary to high school grades. The maximum score is 10.
}
\label{table: comparison_score_level}
\vspace{-2mm}
\setlength{\tabcolsep}{2.5mm}{
\begin{tabular}{l|c|ccc ccccc cccc}
\toprule
Models & Overall & LV1 & LV2 & LV3 & LV4 & LV5 & LV6 & LV7 & LV8 & LV9 & LV10 & LV11 & LV12  \\
\midrule
\multicolumn{14}{c}{Open-source LMMs}\\
\midrule
CogVLM2  & 2.82 & 2.96 & 3.66 & 3.52 & 2.99 & 2.29 & 2.90 & 2.46 & 3.05 & 2.37 & 2.43 & 2.45 & 2.43\\
InternLM-VL & \second{4.48} & \second{4.05} & \best{5.54} & \second{5.27} & \second{4.90} & \second{4.20} & \second{5.12} & \second{4.00} & 4.03 & 2.60 & 4.13 & \second{4.41} & 4.35\\
Qwen2-VL-Instruct & \best{5.26} & \best{4.23}& \best{5.54}& \best{6.07}& \best{5.40}& \best{5.18}& \best{6.05}& \best{5.51}& \second{5.53}& 4.21& \best{4.99}& \best{4.94}& \best{4.64}\\
LLaVA-v1.5 & 2.56 & 2.51& 2.80& 2.96& 2.76& 2.50& 2.37& 2.66& 2.66& 2.37& 2.43& 2.27& 2.34\\
LLaVA-v1.6-mistral & 2.81 & 2.55& 2.88& 2.94& 2.57& 2.43& 2.42& 2.91& 3.36& 3.85& 3.04& 2.68& 2.59\\
CogVLM2 (3-Shot)  &  2.72 & 2.73 & 3.53 & 3.18 & 2.78 & 2.30 & 2.66 & 2.35 & 2.97 & 2.31 & 2.46 & 2.48 & 2.74\\
InternLM-VL (3-Shot)& 4.35 & 3.95 & \second{5.46} & 5.06 & 4.62 & 3.77 & 4.78 & 3.96 & 4.05 & 3.08 & 4.11 & 4.31 & \second{4.36}\\
Qwen2-VL-Instruct (3-Shot) & 4.09 & 3.94& 4.99& 4.94& 4.28& 3.82& 4.73& 3.75& 4.06& 3.28& 3.61& 3.49& 3.47\\

LLaVA-v1.5 (3-Shot) &  3.34 & 2.75& 2.75& 3.10& 2.70& 3.14& 2.49& 3.77& 4.89& \second{5.79}& 3.75& 2.88& 3.42\\
LLaVA-v1.6-mistral (3-Shot) &  3.78 & 3.27& 3.28& 3.50& 3.30& 3.86& 2.62& 3.78& \best{5.67}& \best{6.23}& \second{4.31}& 3.05& 3.73\\

\midrule
\texttt{\textbf{Math-LMM}} (Ours 7B) & 2.46 & 2.27 & 3.17 & 2.95 & 2.97 & 2.40 & 1.83 & 2.16 & 2.67 & 2.16 & 2.13 & 2.26 & 2.11 \\
\texttt{\textbf{Math-LMM}} (Ours 72B) & 4.04 & 3.61& 4.66& 4.52& 3.71& 3.97& 4.49& 3.81& 3.84& 3.64& 4.00& 3.93& 3.86\\
\midrule
\multicolumn{14}{c}{Closed-source LMMs}\\
\midrule
Qwen-VL-Max & 6.50 & 5.91 & 7.30 & 7.54 & 7.11 & 6.61 & 7.02 & 6.1 & 5.91 & 4.73 & 6.01 & 6.18 & 6.17\\
Gemini & 6.02 & 6.30 & 6.85 & 6.83 & 6.46 & 5.75 & 6.40 & 5.67 & 6.46 & 5.01 & 5.42 & 4.96 & 5.62\\
GPT-4o & 7.94 & 7.44 & 8.70 & 8.76 & 8.34 & 8.04 & 8.30 & 7.83 & 7.65 & 6.44 & 7.67 & 7.48 & 7.74\\
Qwen-VL-Max (3-Shot) &  6.21 & 5.66 & 6.70 & 6.43 & 6.18 & 6.12 & 6.91 & 6.00 & 5.72 & 5.50 & 5.67 & 6.57 & 6.51\\
Gemini (3-Shot) &5.89 & 5.83 & 6.54 & 6.47 & 6.06 & 5.44 & 6.26 & 5.51 & 6.59 & 5.36 & 5.77 & 5.21 & 5.44\\
GPT-4o (3-Shot) &  7.85 & 7.27 & 8.53 & 8.63 & 8.45 & 8.09 & 8.36 & 7.87 & 7.65 & 6.06 & 7.48 & 7.22 & 7.46\\
\midrule
Mean accuracy of LMMs &4.62 & 4.29 & 5.16 & 5.15 & 4.75 & 4.44 & 4.76 & 4.45 & 4.82 & 4.17 & 4.41 & 4.26 & 4.39 \\
\bottomrule
\end{tabular}}
\vspace{-2mm}
\end{table*}

\begin{table*}[t!]
\centering
\small
\caption{Comparison of model performances in \textbf{GPT-4o score} across various mathematical subjects. 
}
\label{table: comparison_score_subject}
\vspace{-2mm}
\setlength{\tabcolsep}{1.5mm}{
\begin{tabular}{l|c|ccc ccccc ccccc}
\toprule
Models & Overall & Alg & AnaG & Ari & CombG & Comb & Cnt & Desc & GrphT & Log & MetG & SolG & Stat & TransG \\
\midrule
\multicolumn{14}{c}{Open-source LMMs}\\
\midrule
CogVLM2  & 2.82 & 2.60 & 2.44 & 3.21 & 2.81 & 2.44 & 2.29 & 2.05 & 2.41 & 2.89 & 2.91 & 2.41 & 2.71 & 3.36\\
InternLM-VL & \second{4.48} & 4.51 & 4.02 & \second{5.24} & 2.69 & \second{4.28} & \second{4.08} & 2.11 & \second{3.82} & \second{4.75} & 4.16 & 3.78 & 5.14 & 3.07\\
Qwen2-VL-Instruct &  \best{5.26} & \best{5.73}& \best{4.77}& \best{6.07}& 3.81& \best{4.30}& \best{4.87}& 3.69& 2.94& \best{4.92}& \second{4.71}& \second{4.38}& \best{7.43}& 4.00\\
LLaVA-1.5 &  2.56 & 2.46& 2.38& 2.72& 2.45& 2.30& 2.56& 2.51& 2.35& 2.31& 2.37& 2.63& 3.43& 2.90\\
LLaVA-v1.6-mistral & 2.81 & 2.47& 3.35& 2.68& 2.73& 2.60& 3.09& 3.57& 3.71& 2.52& 2.93& 2.61& 3.86& 3.74\\
CogVLM2 (3-Shot)  & 2.72 & 2.53 & 2.62 & 2.94 & 2.42 & 2.59 & 2.65 & 1.90 & 2.12 & 2.84 & 2.68 & 2.51 & 2.57 & 3.14\\
InternLM-VL (3-Shot)&  4.35 & \second{4.58} & \second{4.14} & 5.08 & 2.91 & 4.25 & 3.73 & 2.06 & 3.06 & 4.44 & 3.91 & 3.65 & 4.71 & 2.93\\
Qwen2-VL-Instruct (3-Shot) & 4.09 & 3.83& 3.67& 4.91& 3.61& 3.34& 3.60& 3.26& 2.82& 3.53& 3.75& 3.23& 5.43& 2.95\\
LLaVA-v1.5 (3-Shot) & 3.34 & 2.91& 3.92& 2.74& \second{5.36}& 2.76& 3.23& \second{5.27}& \second{3.82}& 2.95& 4.04& 3.86& 3.43& \best{4.90}\\
LLaVA-v1.6-mistral (3-Shot) & 3.78 & 3.27& 4.07& 2.94& \best{5.90}& 3.08& 3.51& \best{6.24}& \best{4.82}& 3.27& \best{5.05}& \best{5.02}& 2.43& \second{4.83}\\

\midrule
\texttt{\textbf{Math-LMM}} (Ours 7B) & 2.46 & 2.19 & 2.23 & 2.76 & 2.40 & 1.90 & 1.84 & 2.41 & 2.06 & 2.62 & 2.40 & 2.53 & 1.43 & 2.43\\
\texttt{\textbf{Math-LMM}} (Ours 72B) &  4.04 & 4.25& 3.90& 4.52& 2.80& 3.59& 3.56& 3.26& 2.65& 4.44& 3.74& 3.46& \second{7.00}& 3.24\\

\midrule
\multicolumn{14}{c}{Closed-source LMMs}\\
\midrule
Qwen-VL-Max & 6.50 & 6.66 & 5.80 & 7.41 & 4.30 & 6.07 & 5.84 & 4.65 & 3.82 & 6.16 & 6.12 & 5.86 & 6.29 & 5.07\\
Gemini & 6.02 & 5.65 & 5.29 & 6.72 & 5.72 & 5.15 & 5.32 & 5.17 & 5.47 & 5.30 & 6.06 & 5.71 & 6.00 & 5.05\\
GPT-4o & 7.94 & 8.15 & 7.28 & 8.65 & 6.23 & 7.89 & 7.70 & 6.22 & 7.41 & 7.56 & 7.65 & 7.20 & 8.00 & 7.14\\
Qwen-VL-Max (3-Shot) & 6.21 & 6.52 & 6.14 & 6.73 & 4.74 & 5.44 & 5.45 & 5.74 & 5.59 & 6.06 & 5.96 & 5.81 & 4.86 & 4.81\\
Gemini (3-Shot) &5.89 & 5.72 & 5.41 & 6.45 & 5.62 & 5.49 & 5.36 & 5.34 & 6.18 & 5.33 & 5.78 & 5.41 & 6.14 & 5.02\\
GPT-4o (3-Shot) & 7.85 & 7.99 & 7.07 & 8.61 & 6.12 & 7.74 & 7.51 & 6.31 & 8.24 & 7.52 & 7.46 & 7.22 & 8.57 & 6.90\\
\midrule
Mean accuracy of LMMs &4.62 & 4.56 & 4.36 & 5.02 & 4.03 & 4.18 & 4.23 & 3.99 & 4.07 & 4.41 & 4.54 & 4.29 & 4.97 & 4.19  \\
\bottomrule
\end{tabular}}
\vspace{-2mm}
\end{table*}

\begin{table*}[t!]
\vspace{-2mm}
\centering
 \small
     \caption{Accuracy scores on the \textit{testmini} subset of MathVista.  FQA: figure question answering, GPS: geometry problem solving, MWP: math word problem, TQA: textbook question answering, VQA: visual question answering. Mathematical reasoning types: ALG: algebraic reasoning, ARI: arithmetic reasoning, GEO: geometry reasoning, LOG: logical reasoning, NUM: numeric commonsense, SCI: scientific reasoning, STA: statistical reasoning. 
    }
\label{tab:mathvista}
\vspace{-2mm}
 \setlength{\tabcolsep}{2.0mm}{
    \begin{tabular}{l|c|ccccc|ccccccc}
    \toprule
    Model & Overall & FQA & GPS & MWP & TQA & VQA & ALG & ARI & GEO & LOG & NUM & SCI & STA  \\ 
          \midrule
    \multicolumn{14}{c}{Open-source LMMs} \\
    \midrule
    IDEFICS-9B-Instruct  & 19.8 & 21.6 & 21.1 & 6.5 & 25.9 & 24.0 & 22.1 & 15.0 & 19.8 & \second{18.9} & 9.9 & 24.6 & 18.1 \\
    mPLUG-Owl-LLaMA-7B  & 22.2 & 22.7 & 23.6 & 10.2 & 27.2 & 27.9 & 23.6 & 19.2 & 23.9 & 13.5 & 12.7 & 26.3 & 21.4 \\
    miniGPT4-LLaMA-2-7B  & 23.1 & 18.6 & 26.0 & 13.4 & 30.4 & 30.2 & 28.1 & 21.0 & 24.7 & 16.2 & 16.7 & 25.4 & 17.9 \\
    LLaMA-Adapter-V2-7B  & 23.9 & 21.2 & 25.5 & 11.3 & 32.3 & \second{31.8} & 26.3 & 20.4 & 24.3 & \best{24.3} & 13.9 & 29.5 & 18.3 \\
    LLaVAR  & 25.2 & 21.9 & 25.0 & 16.7 & \second{34.8} & 30.7 & 24.2 & 22.1 & 23.0 & 13.5 & 15.3 & \best{42.6} & 21.9 \\
    InstructBLIP-Vicuna-7B   & 25.3 & 23.1 & 20.7 & \second{18.3} & 32.3 & \best{35.2} & 21.8 & 27.1 & 20.7 & \second{18.9} & 20.4 & 33.0 & 23.1 \\
    LLaVA-LLaMA-2-13B  & 26.1 & \second{26.8} & 29.3 & 16.1 & 32.3 & 26.3 & 27.3 & 20.1 & 28.8 & \best{24.3} & 18.3 & 37.3 & 25.1 \\
    \texttt{\textbf{Math-LMM}} (Ours 7B)  & \second{34.9} & 25.3 & \best{46.6} & \best{46.2} & 34.2 & 24.6 & \best{43.1} & \second{30.3} & \best{45.6} & 16.2 & \second{27.1} & 28.6 & \second{25.9} \\
    \texttt{\textbf{Math-LMM}} (Ours 72B)  & \best{36.3} & \best{36.4} & \second{33.7} & \best{46.2} & \best{36.1} & 29.1 & \second{29.9} & \best{34.6} & \second{33.5} & \best{24.3} & \best{28.5} & \second{41.8} & \best{40.2} \\
    \midrule
    \multicolumn{14}{c}{Closed-source LMMs} \\
    \midrule
    Multimodal Bard  & 34.8 & 26.0 & 47.1 & 29.6 & 48.7 & 26.8 & 46.5 & 28.6 & 47.8 & 13.5 & 14.9 & 47.5 & 33.0 \\
    GPT-4V (Playground)   & 49.9 & 43.1 & 50.5 & 57.5 & 65.2 & 38.0 & 53.0 & 49.0 & 51.0 & 21.6 & 20.1 & 63.1 & 55.8 \\
    \midrule
    \multicolumn{14}{c}{Human Performance} \\
    \midrule
    Human performance  & 60.3 & 59.7 & 48.4 & 73.0 & 63.2 & 55.9 & 50.9 & 59.2 & 51.4 & 40.7 & 53.8 & 64.9 & 63.9 \\
    \bottomrule
       \end{tabular}
    }
\vspace{-2mm}
\end{table*}

\vspace{-2mm}
\section{Experiments}
\vspace{-1mm}
\subsection{Experimental Setups}

\paragraph{Selected LMMs}

We evaluate a series of LLMs on CMM-Math, including the current state-of-the-art open-source and closed-source models. Particularly, we select InternLM-XComposer2.5-VL (InternLM-VL) \cite{internlmxcomposer2_5}, Qwen2-VL-Instruct \cite{Qwen2VL} and CogVLM2-llama3-Chinese-chat (CogVLM2) \cite{wang2023cogvlm} for open-source models, and for closed-source models, we employ Qwen-VL \cite{qwen}, Gemini \cite{geminiteam2024geminifamilyhighlycapable} and GPT-4o \cite{openai2023gpt4}.

\vspace{-2mm}
\paragraph{Datasets and Evaluation Metrics.}
First, we evaluate the performance of typical LMMs on CMM-Math. Then, we conduct experiments to verify the effectiveness of \texttt{MATH-LMM} over MATHVISTA \cite{lu2024mathvista} and MATH-V \cite{wang2024measuring} datasets.
In this paper, we employ the Accuracy and GPT-4o scores to measure the performance of our model and baselines. 
Particularly, we use accuracy for choice and yes-no problems, and use the GPT-4o score for fill-in-the-blank, and analysis problems.
For the GPT-4o score, we use GPT-4o to calculate the scores of generated answers by giving the problems, solutions, answers, and generated responses.
Specially, the prompt we designed is shown in Figure 3. 

\vspace{-2mm}
\paragraph{Implementation Details.}
We employ a lightweight DFN5B-H-14+ \cite{fang2023data} as the vision encoder. This model is trained on a large amount of high-quality paired image-text data using Contrastive Language Image Pre-training (CLIP) and demonstrates superior performance at a comparable scale of parameters. The image encoding module contains only 0.6 billion parameters, which enables our 7B model to perform inference deployment on a single 4090 GPU. In addition, we select an open-source, high-performance Qwen2 \cite{qwen} as our large language model module.

\subsection{Evaluating SOTA LMMs on CMM-Math}

We evaluate the performance of several SOTA LMMs over CMM-Math in terms of accuracy (Table \ref{table: comparison_accuracy_level} and Table \ref{table: comparison_accuracy_subject}) and GPT-4o score (Table \ref{table: comparison_score_level} and Table \ref{table: comparison_score_subject}). 
We also report the performance of various subjects and levels to analyze the LMMs' abilities of multimodal mathematical reasoning.
We give the analysis in two parts: attention to CMM-Math and attention to SOTA LMMs.

\paragraph{\textbf{Attention to CMM-Math}}

We analyze CMM-Math from five perspectives: challenges of CMM-Math, comparisons across different subjects and levels, disparity between accuracy and GPT-4o score, and zero-shot versus few-shot scenarios.

\textit{Challenges of CMM-Math.} From the last row of Table \ref{table: comparison_accuracy_level} and Table \ref{table: comparison_accuracy_subject}, we can observe that the average results across all models are only 35.66. In some subjects, the average results are even below 30, such as combinatorial, descriptive, solid geometry, and graph theory. Meanwhile, GPT-4o (3-Shot) achieves the best result with a score of 65.98 among all models. However, GPT-4o (3-Shot) still performs poorly on difficult levels, like Levels 9 and 10, and faces challenges in certain subjects, such as combinatorial, descriptive, and solid geometry. This indicates that achieving good results on CMM-Math is challenging for most current LMMs, and even the best LMMs struggle to perform exceptionally well on CMM-Math.

\textit{Comparisons Across Different Levels.} Most LMMs perform well on primary school-level problems, but their performance declines at the high school level. As indicated in the last row of Table \ref{table: comparison_accuracy_level} and Table \ref{table: comparison_score_level}, the mean accuracy of LMMs shows an overall downward trend. Specifically, for instance, when employing three-shot prompting with GPT-4o, the model achieves an accuracy of 84.08 on level 2 but only 52.23 on level 10. This phenomenon can also be more intuitively observed in Figure \ref{fig:performance_level}, where for most models, the shaded area representing lower grade levels in the upper semicircle is significantly larger than the area representing higher grade levels in the lower semicircle.

\textit{Comparisons Across Different Subjects.} Most LMMs excel in arithmetic and statistics but show limited proficiency in geometry. We can observe that the mean accuracy of all LMMs across all subjects is 35.66, with LMMs achieving an accuracy exceeding 40 in arithmetic and statistics subjects. In contrast, in geometry, including analytical, combinatorial, descriptive, metric, and transformation geometry, the accuracy of LMMs is below 35.66, with the worst performance at 23.17 for descriptive geometry.

\begin{table*}[t!]
\vspace{-2mm}
\centering
\small
\caption{Comparison of model performances on MATH-V across various mathematical subjects. Alg: algebra, AnaG: analytic geometry, Ari: arithmetic, CombG: combinatorial geometry, Comb: combinatorics, Cnt: counting, DescG: descriptive geometry, GrphT: graph theory, Log: logic, Angle: metric geometry - angle, Area: metric geometry - area, Len: metric geometry-length, SolG: solid geometry, Stat: statistics, Topo: topology, TransG: transformation geometry. 
}
\label{tab:math-v}
\vspace{-2mm}
\setlength{\tabcolsep}{1.0mm}{
\begin{tabular}{l|c|cccccccccccccccc}
\toprule
Model & Overall & Alg & AnaG & Ari & CombG & Comb & Cnt & DescG & GrphT & Log & Angle & Area & Len & SolG & Stat & Topo & TransG \\
\toprule

\multicolumn{18}{c}{Open-source LMMs}\\
\midrule
LLaVA-v1.5-7B & 8.52 & 7.00 & 7.10 & 10.70 & 7.10 & 4.80 & 10.50 & 7.70 & 10.00 & 9.20 & 15.60 & 10.20 & 9.80 & 5.30 & 8.60 & 4.40 & 4.80 \\
SPHINX (V2) & 9.70 & 6.70 & 7.10 & 12.90 & 7.50 & 7.70 & 6.00 & 9.60 & \second{16.70} & 10.10 & 11.00 & 11.80 & 12.50 & 8.20 & 8.60 & 8.70 & 6.00 \\
ShareGPT4V-7B & 10.53 & 5.50 & 3.60 & 12.90 & 10.10 & 4.80 & 7.50 & 11.50 & 14.40 & 10.90 & 16.20 & 11.80 & 12.30 & 9.80 & \second{15.50} & \second{17.40} & 11.30 \\
LLaVA-v1.5-13B & 11.12 & 7.00 & 14.30 & 14.30 & 9.10 & 6.60 & 6.00 & 13.50 & 5.60 & 13.50 & 10.40 & 12.60 & 14.70 & 11.50 & 13.80 & 13.00 & 10.70 \\
ShareGPT4V-13B & 11.88 & 7.50 & 15.50 & \best{16.40} & 10.70 & 8.90 & 9.00 & 11.50 & 8.90 & 7.60 & 11.60 & 13.00 & 17.40 & 10.30 & 8.60 & 8.70 & 12.50 \\
SPHINX-MoE & 14.18 & 7.80 & \second{17.90} & 14.30 & \second{15.60} & \second{9.50} & \best{11.90} & 12.50 & 15.6 & 12.60 & 16.20 & 15.60 & \second{17.80} & \second{13.50} & 12.10 & 8.70 & 16.10 \\
InternLM-VL & \second{14.54} & \second{9.30} & 15.50 & 12.10 & 15.30 & \best{11.30} & \second{10.50} & 14.40 & \best{22.20} & \second{19.30} & \second{19.70} & \second{15.6} & 15.00 & 11.90 & \second{15.50} & \best{26.10} & 15.50 \\
\texttt{\textbf{Math-LMM}} (Ours 7B) & 11.58  & 7.30  & 8.30  & 10.70 & 14.00  & 7.10 & 7.40  & \best{16.40}  & 12.20  &  9.20 & 14.50  & 10.60 & 14.90 & 9.00 & 8.60 & \best{26.10} & \second{16.70}  \\
\texttt{\textbf{Math-LMM}} (Ours 72B) & \best{17.53}  & \best{10.70}  & \best{28.60}  & \second{15.00} & \best{20.10}  & \best{11.30} & \best{11.90}  & \second{15.40}  & \second{16.70}  & \best{21.00} & \best{22.50}  & \best{18.40} & \best{20.00} & \best{15.60} & \best{20.70} & 8.70 & \best{19.60}  \\

\midrule
\multicolumn{18}{c}{Closed-source LMMs}\\
\midrule
Qwen-VL-Plus & 10.72 & 11.30 & 17.90 & 14.30 & 12.70 & 4.80 & 10.50 & 15.40 & 8.90 & 14.30 & 11.60 & 6.40 & 10.00 & 14.30 & 6.90 & 8.70 & 11.31 \\
Qwen-VL-Max & 15.59 & 10.70 & 19.10 & 20.00 & 16.90 & 12.50 & 17.90 & 16.40 & 12.20 & 21.00 & 13.30 & 14.20 & 19.80 & 11.50 & 20.70 & 13.00 & 17.30 \\
Gemini Pro & 17.66 & 15.10 & 10.70 & 20.70 & 20.10 & 11.90 & 7.50 & 20.20 & 21.10 & 16.80 & 19.10 & 19.00 & 20.00 & 14.30 & 13.80 & 17.40 & 20.80 \\
GPT-4V & 22.76 & 27.30 & 32.10 & 35.70 & 21.10 & 16.70 & 13.40 & 22.10 & 14.40 & 16.80 & 22.00 & 22.20 & 20.90 & 23.80 & 24.10 & 21.70 & 25.60 \\

\midrule
\multicolumn{18}{c}{Human Performance}\\
\midrule
Human (testmini) & 75.66 & 57.90 & 79.00 & 100.00 & 100.00 & 47.40 & 94.70 & 89.50 & 63.20 & 63.20 & 36.80 & 52.60 & 73.70 & 89.50 & 89.50 & 100.00 & 73.70 \\

\bottomrule
\end{tabular}%
}
\vspace{-2mm}
\end{table*}

\textit{Disparity Between Accuracy and GPT-4o Score.} 
When comparing the performance of LMMs in both accuracy and GPT-4o score, we observe that closed-source models maintain consistent performance while open-source models show inconsistencies. Specifically, GPT-4o consistently ranks highest, while Gemini ranks lowest. 
CogVLM2 has higher accuracy than InternLM-VL but is lower on the GPT-4o score. This discrepancy may arise from differences in the training data distribution, where the datasets might focus more on choice and yes-no questions while lacking materials for problem-solving and analytical tasks. 
Moreover, this inconsistency could also result from insufficient training data, leading to models that obtain correct answers but struggle to provide high-quality analytical processes. 
This highlights that our CMM-Math, unlike other math test sets that offer only choice questions, provides a more comprehensive evaluation of models' analytical and problem-solving abilities.

\textit{Zero-shot versus 3-shot Scenarios.} 
All few-shot models perform better than zero-shot prompting in terms of accuracy. This indicates that few-shot prompting helps models obtain correct answers. Conversely, regarding the GPT-4o score, most models perform better with zero-shot prompting than with few-shot prompting, except for the LLaVA-v1.5 and LLaVA-v1.6-mistral models. This suggests that few-shot prompting does not necessarily lead to higher quality outputs and analyses. Our view is that the content of few-shot prompts may influence how models conduct their own analysis and reasoning, potentially leading to poorer performance.
\paragraph{\textbf{Attention to SOTA LMMs}}

We separately provide an analysis of the performance of closed-source and open-source LMMs.

\textit{Closed-source LMMs.} 
We observe that all closed-source models significantly outperform open-source models. Both in terms of accuracy and GPT-4o score, GPT-4o and Qwen-VL demonstrate the best performance on the CMM-Math in most cases. However, we note that GPT-4o performs optimally on simple and moderately difficult levels, while Qwen-VL excels on the most challenging levels, including levels 10, 11, and 12. This suggests that Qwen-VL may have a stronger advantage in handling more complex problems.

\textit{Open-source LMMs.} 
In terms of accuracy, our \texttt{Math-LMM} (72B) and Qwen2-VL-Instruct (3-Shot) achieve the best and second-best performances, with scores of 48.57 and 46.29, respectively. On the most challenging levels 10, 11, and 12, \texttt{Math-LMM} (72B) achieves the best results, indicating that \texttt{Math-LMM} has a stronger advantage in handling more complex problems. Furthermore, regarding the GPT-4o score, we notice that \texttt{Math-LMM} achieved only ordinary results, while Qwen2-VL-Instruct obtains the best result of 5.26. This could be because the training data for \texttt{Math-LMM} is still limited, leading to weaker language expression and analytical reasoning capabilities. 

\subsection{Evaluating \texttt{Math-LMM}}
To evaluate the effectiveness of our \texttt{Math-LMM}, we conduct experiments on MATHVISTA \cite{lu2024mathvista} and MATH-V \cite{wang2024measuring} (Table \ref{tab:mathvista} and Table \ref{tab:math-v}).
These two datasets are widely used to evaluate the capability of large multimodal models in solving English mathematical problems.

\vspace{-2mm}
\paragraph{\textbf{Main Results over MATHVISTA}}
We select LMMs used in the MathVista \cite{lu2024mathvista} experiments for comparison with \texttt{Math-LMM}, including IDEFICS-9B-Instruct \cite{laurençon2023obelicsopenwebscalefiltered}, mPLUG-Owl-LLaMA-7B \cite{ye2024mplugowlmodularizationempowerslarge}, miniGPT4-LLaMA-2-7B \cite{zhu2023minigpt}, LLaMA-Adapter-V2-7B \cite{gao2023llamaadapterv2parameterefficientvisual}, LLaVAR \cite{zhang2024llavarenhancedvisualinstruction}, InstructBLIP-Vicuna-7B \cite{dai2023instructblipgeneralpurposevisionlanguagemodels}, LLaVA-LLaMA-2-13B \cite{liu2024visual}, Multimodal Bard \cite{google2023bard}, and GPT-4V (Playground) \cite{openai2023gpt4}.

In the results of open-source models shown in Table \ref{tab:mathvista}, our \texttt{Math-LMM} achieves the best (36.3) and suboptimal (34.9) results with 72B and 7B versions. Across five problem tasks, \texttt{Math-LMM} (7B) obtains the best performance on two tasks and \texttt{Math-LMM} (72B) achieves the best performance on three tasks and the second-best on one task. Furthermore, our \texttt{Math-LMM} almost achieves the best performance across all types of mathematical reasoning, except for scientific reasoning, where it achieved a suboptimal performance. Additionally, in numeric commonsense problems, both parameter versions of our \texttt{Math-LMM} perform exceptionally well, even surpassing closed-source models such as Multimodal Bard \cite{google2023bard} and GPT-4V (Playground). These outcomes suggest that \texttt{Math-LMM} demonstrates commendable capabilities in addressing English multimodal mathematical questions. Detailed comparative analyses of other models can be found in MathVista \cite{lu2024mathvista}.

\vspace{-2mm}
\paragraph{\textbf{Main Results over MATH-V}}
We also compare our \texttt{Math-LMM} with existing LLMs over MATH-V \cite{wang2024measuring}, including Qwen-VL-Plus \cite{bai2023qwenvlversatilevisionlanguagemodel}, Qwen-VL-Max \cite{bai2023qwenvlversatilevisionlanguagemodel}, Gemini Pro \cite{geminiteam2024geminifamilyhighlycapable}, GPT-4V \cite{openai2023gpt4}, LLaVa-v1.5-7B \cite{liu2024visual}, SPHINX \cite{lin2023sphinxjointmixingweights}, ShareGPT-4V-7B \cite{chen2023sharegpt4vimprovinglargemultimodal}, LLaVa-v1.5-13B \cite{liu2024visual}, ShareGPT-4V-13B \cite{chen2023sharegpt4vimprovinglargemultimodal}, InternLM-XComposer2-VL \cite{internlmxcomposer2}, and SPHINX-MoE \cite{lin2023sphinxjointmixingweights}.

In Table \ref{tab:math-v}, our \texttt{Math-LMM} (72B) achieves the best performance among in open-source models. Although the \texttt{Math-LMM} (7B) does not achieve the second-best performance, compared to other open-source models of the same parameter scale, such as LLaVa-v1.5-7B \cite{liu2024visual}, SPHINX \cite{lin2023sphinxjointmixingweights}, and ShareGPT-4V-7B \cite{chen2023sharegpt4vimprovinglargemultimodal}, \texttt{Math-LMM} (7B) still achieves the best performance. This may be because MATH-V \cite{wang2024measuring} is a more challenging test set than MATHVISTA \cite{lu2024mathvista}, where model parameter scale has a greater impact on performance. Meanwhile, among open-source models, \texttt{Math-LMM} almost achieves the best performance across all 16 subjects, except for the subjects of arithmetic and graph theory. Notably, compared to closed-source models, \texttt{Math-LMM} also achieves the best performance among all models on the subjects of logic, metric geometry-angle, and topology. These results indicate that \texttt{Math-LMM} also has a certain level of competitiveness on harder English multimodal mathematical problems.

\vspace{-2mm}
\section{Conclusion}
In this paper, we introduce CMM-Math, a comprehensive Chinese multimodal mathematical dataset designed to evaluate and enhance the performance of LMMs in mathematical reasoning. CMM-Math is distinguished by its scale, diversity, and complexity, comprising over 28,000 high-quality samples that span 12 grade levels and include various problem types. This dataset serves as both a benchmark and a training resource, addressing the gap in non-English, specifically Chinese, multimodal mathematical datasets.
Our experiments reveal that existing state-of-the-art LMMs struggle with the challenges posed by the CMM-Math dataset, highlighting the need for further advancements in this field. We also propose a new math-specific LMM, \texttt{Math-LMM}, which is trained through a three-stage process: foundational pre-training, foundational fine-tuning, and mathematical fine-tuning. The results from our evaluations demonstrate that \texttt{Math-LMM} improves performance in multimodal mathematical reasoning by comparing with open-source LMMs.


\bibliographystyle{ACM-Reference-Format}
\bibliography{custom}


\begin{thebibliography}{42}


\ifx \showCODEN    \undefined \def \showCODEN     #1{\unskip}     \fi
\ifx \showDOI      \undefined \def \showDOI       #1{#1}\fi
\ifx \showISBNx    \undefined \def \showISBNx     #1{\unskip}     \fi
\ifx \showISBNxiii \undefined \def \showISBNxiii  #1{\unskip}     \fi
\ifx \showISSN     \undefined \def \showISSN      #1{\unskip}     \fi
\ifx \showLCCN     \undefined \def \showLCCN      #1{\unskip}     \fi
\ifx \shownote     \undefined \def \shownote      #1{#1}          \fi
\ifx \showarticletitle \undefined \def \showarticletitle #1{#1}   \fi
\ifx \showURL      \undefined \def \showURL       {\relax}        \fi
\providecommand\bibfield[2]{#2}
\providecommand\bibinfo[2]{#2}
\providecommand\natexlab[1]{#1}
\providecommand\showeprint[2][]{arXiv:#2}

\bibitem[Amini et~al\mbox{.}(2019)]%
        {amini2019mathqa}
\bibfield{author}{\bibinfo{person}{Aida Amini}, \bibinfo{person}{Saadia Gabriel}, \bibinfo{person}{Shanchuan Lin}, \bibinfo{person}{Rik Koncel-Kedziorski}, \bibinfo{person}{Yejin Choi}, {and} \bibinfo{person}{Hannaneh Hajishirzi}.} \bibinfo{year}{2019}\natexlab{}.
\newblock \showarticletitle{MathQA: Towards Interpretable Math Word Problem Solving with Operation-Based Formalisms}. In \bibinfo{booktitle}{\emph{NAACL}}. \bibinfo{pages}{2357--2367}.
\newblock


\bibitem[Azerbayev et~al\mbox{.}(2023)]%
        {azerbayev8llemma}
\bibfield{author}{\bibinfo{person}{Zhangir Azerbayev}, \bibinfo{person}{Hailey Schoelkopf}, \bibinfo{person}{Keiran Paster}, \bibinfo{person}{Marco Dos~Santos}, \bibinfo{person}{Stephen McAleer}, \bibinfo{person}{Albert~Q Jiang}, \bibinfo{person}{Jia Deng}, \bibinfo{person}{Stella Biderman}, {and} \bibinfo{person}{Sean Welleck}.} \bibinfo{year}{2023}\natexlab{}.
\newblock \showarticletitle{LLEMMA: AN OPEN LANGUAGE MODEL FOR MATHEMATICS}.
\newblock \bibinfo{journal}{\emph{Minerva}}  \bibinfo{volume}{8} (\bibinfo{year}{2023}), \bibinfo{pages}{164B}.
\newblock


\bibitem[Bai et~al\mbox{.}(2023a)]%
        {qwen}
\bibfield{author}{\bibinfo{person}{Jinze Bai}, \bibinfo{person}{Shuai Bai}, \bibinfo{person}{Yunfei Chu}, {and} \bibinfo{person}{Zeyu Cui}.} \bibinfo{year}{2023}\natexlab{a}.
\newblock \showarticletitle{Qwen Technical Report}.
\newblock \bibinfo{journal}{\emph{arXiv preprint arXiv:2309.16609}} (\bibinfo{year}{2023}).
\newblock


\bibitem[Bai et~al\mbox{.}(2023b)]%
        {bai2023qwenvlversatilevisionlanguagemodel}
\bibfield{author}{\bibinfo{person}{Jinze Bai}, \bibinfo{person}{Shuai Bai}, \bibinfo{person}{Shusheng Yang}, \bibinfo{person}{Shijie Wang}, \bibinfo{person}{Sinan Tan}, \bibinfo{person}{Peng Wang}, \bibinfo{person}{Junyang Lin}, \bibinfo{person}{Chang Zhou}, {and} \bibinfo{person}{Jingren Zhou}.} \bibinfo{year}{2023}\natexlab{b}.
\newblock \bibinfo{title}{Qwen-VL: A Versatile Vision-Language Model for Understanding, Localization, Text Reading, and Beyond}.
\newblock
\newblock
\showeprint[arxiv]{2308.12966}


\bibitem[Chen et~al\mbox{.}(2023)]%
        {chen2023sharegpt4vimprovinglargemultimodal}
\bibfield{author}{\bibinfo{person}{Lin Chen}, \bibinfo{person}{Jinsong Li}, \bibinfo{person}{Xiaoyi Dong}, \bibinfo{person}{Pan Zhang}, \bibinfo{person}{Conghui He}, \bibinfo{person}{Jiaqi Wang}, \bibinfo{person}{Feng Zhao}, {and} \bibinfo{person}{Dahua Lin}.} \bibinfo{year}{2023}\natexlab{}.
\newblock \bibinfo{title}{ShareGPT4V: Improving Large Multi-Modal Models with Better Captions}.
\newblock
\newblock
\showeprint[arxiv]{2311.12793}


\bibitem[Cobbe et~al\mbox{.}(2021)]%
        {cobbe2021training}
\bibfield{author}{\bibinfo{person}{Karl Cobbe}, \bibinfo{person}{Vineet Kosaraju}, \bibinfo{person}{Mohammad Bavarian}, \bibinfo{person}{Mark Chen}, \bibinfo{person}{Heewoo Jun}, \bibinfo{person}{Lukasz Kaiser}, \bibinfo{person}{Matthias Plappert}, \bibinfo{person}{Jerry Tworek}, \bibinfo{person}{Jacob Hilton}, \bibinfo{person}{Reiichiro Nakano}, {et~al\mbox{.}}} \bibinfo{year}{2021}\natexlab{}.
\newblock \showarticletitle{Training verifiers to solve math word problems}.
\newblock \bibinfo{journal}{\emph{arXiv preprint arXiv:2110.14168}} (\bibinfo{year}{2021}).
\newblock


\bibitem[Dai et~al\mbox{.}(2023)]%
        {dai2023instructblipgeneralpurposevisionlanguagemodels}
\bibfield{author}{\bibinfo{person}{Wenliang Dai}, \bibinfo{person}{Junnan Li}, \bibinfo{person}{Dongxu Li}, \bibinfo{person}{Anthony Meng~Huat Tiong}, \bibinfo{person}{Junqi Zhao}, \bibinfo{person}{Weisheng Wang}, \bibinfo{person}{Boyang Li}, \bibinfo{person}{Pascale Fung}, {and} \bibinfo{person}{Steven Hoi}.} \bibinfo{year}{2023}\natexlab{}.
\newblock \bibinfo{title}{InstructBLIP: Towards General-purpose Vision-Language Models with Instruction Tuning}.
\newblock
\newblock
\showeprint[arxiv]{2305.06500}


\bibitem[Dong et~al\mbox{.}(2024)]%
        {internlmxcomposer2}
\bibfield{author}{\bibinfo{person}{Xiaoyi Dong}, \bibinfo{person}{Pan Zhang}, \bibinfo{person}{Yuhang Zang}, \bibinfo{person}{Yuhang Cao}, \bibinfo{person}{Bin Wang}, \bibinfo{person}{Linke Ouyang}, \bibinfo{person}{Xilin Wei}, \bibinfo{person}{Songyang Zhang}, \bibinfo{person}{Haodong Duan}, \bibinfo{person}{Maosong Cao}, \bibinfo{person}{Wenwei Zhang}, \bibinfo{person}{Yining Li}, \bibinfo{person}{Hang Yan}, \bibinfo{person}{Yang Gao}, \bibinfo{person}{Xinyue Zhang}, \bibinfo{person}{Wei Li}, \bibinfo{person}{Jingwen Li}, \bibinfo{person}{Kai Chen}, \bibinfo{person}{Conghui He}, \bibinfo{person}{Xingcheng Zhang}, \bibinfo{person}{Yu Qiao}, \bibinfo{person}{Dahua Lin}, {and} \bibinfo{person}{Jiaqi Wang}.} \bibinfo{year}{2024}\natexlab{}.
\newblock \showarticletitle{InternLM-XComposer2: Mastering Free-form Text-Image Composition and Comprehension in Vision-Language Large Model}.
\newblock \bibinfo{journal}{\emph{arXiv preprint arXiv:2401.16420}} (\bibinfo{year}{2024}).
\newblock


\bibitem[Dosovitskiy et~al\mbox{.}(2021)]%
        {DBLP:conf/iclr/DosovitskiyB0WZ21}
\bibfield{author}{\bibinfo{person}{Alexey Dosovitskiy}, \bibinfo{person}{Lucas Beyer}, \bibinfo{person}{Alexander Kolesnikov}, \bibinfo{person}{Dirk Weissenborn}, \bibinfo{person}{Xiaohua Zhai}, \bibinfo{person}{Thomas Unterthiner}, \bibinfo{person}{Mostafa Dehghani}, \bibinfo{person}{Matthias Minderer}, \bibinfo{person}{Georg Heigold}, \bibinfo{person}{Sylvain Gelly}, \bibinfo{person}{Jakob Uszkoreit}, {and} \bibinfo{person}{Neil Houlsby}.} \bibinfo{year}{2021}\natexlab{}.
\newblock \showarticletitle{An Image is Worth 16x16 Words: Transformers for Image Recognition at Scale}. In \bibinfo{booktitle}{\emph{ICLR}}.
\newblock


\bibitem[Fang et~al\mbox{.}(2023)]%
        {fang2023data}
\bibfield{author}{\bibinfo{person}{Alex Fang}, \bibinfo{person}{Albin~Madappally Jose}, \bibinfo{person}{Amit Jain}, \bibinfo{person}{Ludwig Schmidt}, \bibinfo{person}{Alexander Toshev}, {and} \bibinfo{person}{Vaishaal Shankar}.} \bibinfo{year}{2023}\natexlab{}.
\newblock \showarticletitle{Data filtering networks}.
\newblock \bibinfo{journal}{\emph{arXiv preprint arXiv:2309.17425}} (\bibinfo{year}{2023}).
\newblock


\bibitem[Frieder et~al\mbox{.}(2024)]%
        {frieder2024mathematical}
\bibfield{author}{\bibinfo{person}{Simon Frieder}, \bibinfo{person}{Luca Pinchetti}, \bibinfo{person}{Ryan-Rhys Griffiths}, \bibinfo{person}{Tommaso Salvatori}, \bibinfo{person}{Thomas Lukasiewicz}, \bibinfo{person}{Philipp Petersen}, {and} \bibinfo{person}{Julius Berner}.} \bibinfo{year}{2024}\natexlab{}.
\newblock \showarticletitle{Mathematical capabilities of chatgpt}.
\newblock \bibinfo{journal}{\emph{NeurIPS}}  \bibinfo{volume}{36} (\bibinfo{year}{2024}).
\newblock


\bibitem[Gao et~al\mbox{.}(2023b)]%
        {gao2023pal}
\bibfield{author}{\bibinfo{person}{Luyu Gao}, \bibinfo{person}{Aman Madaan}, \bibinfo{person}{Shuyan Zhou}, \bibinfo{person}{Uri Alon}, \bibinfo{person}{Pengfei Liu}, \bibinfo{person}{Yiming Yang}, \bibinfo{person}{Jamie Callan}, {and} \bibinfo{person}{Graham Neubig}.} \bibinfo{year}{2023}\natexlab{b}.
\newblock \showarticletitle{Pal: Program-aided language models}. In \bibinfo{booktitle}{\emph{ICML}}. \bibinfo{pages}{10764--10799}.
\newblock


\bibitem[Gao et~al\mbox{.}(2023a)]%
        {gao2023llamaadapterv2parameterefficientvisual}
\bibfield{author}{\bibinfo{person}{Peng Gao}, \bibinfo{person}{Jiaming Han}, \bibinfo{person}{Renrui Zhang}, \bibinfo{person}{Ziyi Lin}, \bibinfo{person}{Shijie Geng}, \bibinfo{person}{Aojun Zhou}, \bibinfo{person}{Wei Zhang}, \bibinfo{person}{Pan Lu}, \bibinfo{person}{Conghui He}, \bibinfo{person}{Xiangyu Yue}, \bibinfo{person}{Hongsheng Li}, {and} \bibinfo{person}{Yu Qiao}.} \bibinfo{year}{2023}\natexlab{a}.
\newblock \bibinfo{title}{LLaMA-Adapter V2: Parameter-Efficient Visual Instruction Model}.
\newblock
\newblock
\showeprint[arxiv]{2304.15010}


\bibitem[Gheini et~al\mbox{.}(2021)]%
        {gheini2021cross}
\bibfield{author}{\bibinfo{person}{Mozhdeh Gheini}, \bibinfo{person}{Xiang Ren}, {and} \bibinfo{person}{Jonathan May}.} \bibinfo{year}{2021}\natexlab{}.
\newblock \showarticletitle{Cross-attention is all you need: Adapting pretrained transformers for machine translation}.
\newblock \bibinfo{journal}{\emph{arXiv preprint arXiv:2104.08771}} (\bibinfo{year}{2021}).
\newblock


\bibitem[Google(2023)]%
        {google2023bard}
\bibfield{author}{\bibinfo{person}{Google}.} \bibinfo{year}{2023}\natexlab{}.
\newblock \bibinfo{title}{Bard}.
\newblock
\newblock
\urldef\tempurl%
\url{https://bard.google.com/}
\showURL{%
\tempurl}


\bibitem[Hendrycks et~al\mbox{.}({[n.\,d.]})]%
        {hendrycks2measuring}
\bibfield{author}{\bibinfo{person}{Dan Hendrycks}, \bibinfo{person}{Collin Burns}, \bibinfo{person}{Saurav Kadavath}, \bibinfo{person}{Akul Arora}, \bibinfo{person}{Steven Basart}, \bibinfo{person}{Eric Tang}, \bibinfo{person}{Dawn Song}, {and} \bibinfo{person}{Jacob Steinhardt}.} \bibinfo{year}{[n.\,d.]}\natexlab{}.
\newblock \showarticletitle{Measuring Mathematical Problem Solving With the MATH Dataset}.
\newblock \bibinfo{journal}{\emph{Sort}} \bibinfo{volume}{2}, \bibinfo{number}{4} (\bibinfo{year}{[n.\,d.]}), \bibinfo{pages}{0--6}.
\newblock


\bibitem[Hendrycks et~al\mbox{.}(2021a)]%
        {hendrycks2021measuring}
\bibfield{author}{\bibinfo{person}{Dan Hendrycks}, \bibinfo{person}{Collin Burns}, \bibinfo{person}{Saurav Kadavath}, \bibinfo{person}{Akul Arora}, \bibinfo{person}{Steven Basart}, \bibinfo{person}{Eric Tang}, \bibinfo{person}{Dawn Song}, {and} \bibinfo{person}{Jacob Steinhardt}.} \bibinfo{year}{2021}\natexlab{a}.
\newblock \showarticletitle{Measuring mathematical problem solving with the math dataset}.
\newblock \bibinfo{journal}{\emph{arXiv preprint arXiv:2103.03874}} (\bibinfo{year}{2021}).
\newblock


\bibitem[Hendrycks et~al\mbox{.}(2021b)]%
        {hendrycksmath2021}
\bibfield{author}{\bibinfo{person}{Dan Hendrycks}, \bibinfo{person}{Collin Burns}, \bibinfo{person}{Saurav Kadavath}, \bibinfo{person}{Akul Arora}, \bibinfo{person}{Steven Basart}, \bibinfo{person}{Eric Tang}, \bibinfo{person}{Dawn Song}, {and} \bibinfo{person}{Jacob Steinhardt}.} \bibinfo{year}{2021}\natexlab{b}.
\newblock \showarticletitle{Measuring Mathematical Problem Solving With the MATH Dataset}.
\newblock \bibinfo{journal}{\emph{arXiv preprint arXiv:2103.03874}} (\bibinfo{year}{2021}).
\newblock


\bibitem[Jiang et~al\mbox{.}(2024)]%
        {Jiang2024MANTISIM}
\bibfield{author}{\bibinfo{person}{Dongfu Jiang}, \bibinfo{person}{Xuan He}, \bibinfo{person}{Huaye Zeng}, \bibinfo{person}{Cong Wei}, \bibinfo{person}{Max~W.F. Ku}, \bibinfo{person}{Qian Liu}, {and} \bibinfo{person}{Wenhu Chen}.} \bibinfo{year}{2024}\natexlab{}.
\newblock \showarticletitle{MANTIS: Interleaved Multi-Image Instruction Tuning}. \bibinfo{publisher}{arXiv2405.01483}.
\newblock


\bibitem[Laurençon et~al\mbox{.}(2023)]%
        {laurençon2023obelicsopenwebscalefiltered}
\bibfield{author}{\bibinfo{person}{Hugo Laurençon}, \bibinfo{person}{Lucile Saulnier}, \bibinfo{person}{Léo Tronchon}, \bibinfo{person}{Stas Bekman}, \bibinfo{person}{Amanpreet Singh}, \bibinfo{person}{Anton Lozhkov}, \bibinfo{person}{Thomas Wang}, \bibinfo{person}{Siddharth Karamcheti}, \bibinfo{person}{Alexander~M. Rush}, \bibinfo{person}{Douwe Kiela}, \bibinfo{person}{Matthieu Cord}, {and} \bibinfo{person}{Victor Sanh}.} \bibinfo{year}{2023}\natexlab{}.
\newblock \bibinfo{title}{OBELICS: An Open Web-Scale Filtered Dataset of Interleaved Image-Text Documents}.
\newblock
\newblock
\showeprint[arxiv]{2306.16527}


\bibitem[Li et~al\mbox{.}(2023)]%
        {li2023blip}
\bibfield{author}{\bibinfo{person}{Junnan Li}, \bibinfo{person}{Dongxu Li}, \bibinfo{person}{Silvio Savarese}, {and} \bibinfo{person}{Steven Hoi}.} \bibinfo{year}{2023}\natexlab{}.
\newblock \showarticletitle{Blip-2: Bootstrapping language-image pre-training with frozen image encoders and large language models}. In \bibinfo{booktitle}{\emph{ICML}}. \bibinfo{pages}{19730--19742}.
\newblock


\bibitem[Lin et~al\mbox{.}(2023)]%
        {lin2023sphinxjointmixingweights}
\bibfield{author}{\bibinfo{person}{Ziyi Lin}, \bibinfo{person}{Chris Liu}, \bibinfo{person}{Renrui Zhang}, \bibinfo{person}{Peng Gao}, \bibinfo{person}{Longtian Qiu}, \bibinfo{person}{Han Xiao}, \bibinfo{person}{Han Qiu}, \bibinfo{person}{Chen Lin}, \bibinfo{person}{Wenqi Shao}, \bibinfo{person}{Keqin Chen}, \bibinfo{person}{Jiaming Han}, \bibinfo{person}{Siyuan Huang}, \bibinfo{person}{Yichi Zhang}, \bibinfo{person}{Xuming He}, \bibinfo{person}{Hongsheng Li}, {and} \bibinfo{person}{Yu Qiao}.} \bibinfo{year}{2023}\natexlab{}.
\newblock \bibinfo{title}{SPHINX: The Joint Mixing of Weights, Tasks, and Visual Embeddings for Multi-modal Large Language Models}.
\newblock
\newblock
\showeprint[arxiv]{2311.07575}


\bibitem[Liu et~al\mbox{.}(2024b)]%
        {liu2024visual}
\bibfield{author}{\bibinfo{person}{Haotian Liu}, \bibinfo{person}{Chunyuan Li}, \bibinfo{person}{Qingyang Wu}, {and} \bibinfo{person}{Yong~Jae Lee}.} \bibinfo{year}{2024}\natexlab{b}.
\newblock \showarticletitle{Visual instruction tuning}.
\newblock \bibinfo{journal}{\emph{NeurIPS}}  \bibinfo{volume}{36} (\bibinfo{year}{2024}).
\newblock


\bibitem[Liu et~al\mbox{.}(2023)]%
        {liu2023mathematical}
\bibfield{author}{\bibinfo{person}{Wentao Liu}, \bibinfo{person}{Hanglei Hu}, \bibinfo{person}{Jie Zhou}, \bibinfo{person}{Yuyang Ding}, \bibinfo{person}{Junsong Li}, \bibinfo{person}{Jiayi Zeng}, \bibinfo{person}{Mengliang He}, \bibinfo{person}{Qin Chen}, \bibinfo{person}{Bo Jiang}, \bibinfo{person}{Aimin Zhou}, {et~al\mbox{.}}} \bibinfo{year}{2023}\natexlab{}.
\newblock \showarticletitle{Mathematical language models: A survey}.
\newblock \bibinfo{journal}{\emph{arXiv preprint arXiv:2312.07622}} (\bibinfo{year}{2023}).
\newblock


\bibitem[Liu et~al\mbox{.}(2024a)]%
        {liu2024mmdumultiturnmultiimagedialog}
\bibfield{author}{\bibinfo{person}{Ziyu Liu}, \bibinfo{person}{Tao Chu}, \bibinfo{person}{Yuhang Zang}, \bibinfo{person}{Xilin Wei}, \bibinfo{person}{Xiaoyi Dong}, \bibinfo{person}{Pan Zhang}, \bibinfo{person}{Zijian Liang}, \bibinfo{person}{Yuanjun Xiong}, \bibinfo{person}{Yu Qiao}, \bibinfo{person}{Dahua Lin}, {and} \bibinfo{person}{Jiaqi Wang}.} \bibinfo{year}{2024}\natexlab{a}.
\newblock \bibinfo{title}{MMDU: A Multi-Turn Multi-Image Dialog Understanding Benchmark and Instruction-Tuning Dataset for LVLMs}.
\newblock
\newblock
\showeprint[arxiv]{2406.11833}


\bibitem[Lu et~al\mbox{.}(2024)]%
        {lu2024mathvista}
\bibfield{author}{\bibinfo{person}{Pan Lu}, \bibinfo{person}{Hritik Bansal}, \bibinfo{person}{Tony Xia}, \bibinfo{person}{Jiacheng Liu}, \bibinfo{person}{Chunyuan Li}, \bibinfo{person}{Hannaneh Hajishirzi}, \bibinfo{person}{Hao Cheng}, \bibinfo{person}{Kai-Wei Chang}, \bibinfo{person}{Michel Galley}, {and} \bibinfo{person}{Jianfeng Gao}.} \bibinfo{year}{2024}\natexlab{}.
\newblock \showarticletitle{MathVista: Evaluating Mathematical Reasoning of Foundation Models in Visual Contexts}. In \bibinfo{booktitle}{\emph{ICLR}}.
\newblock


\bibitem[Mishra et~al\mbox{.}(2022)]%
        {mishra2022lila}
\bibfield{author}{\bibinfo{person}{Swaroop Mishra}, \bibinfo{person}{Matthew Finlayson}, \bibinfo{person}{Pan Lu}, \bibinfo{person}{Leonard Tang}, \bibinfo{person}{Sean Welleck}, \bibinfo{person}{Chitta Baral}, \bibinfo{person}{Tanmay Rajpurohit}, \bibinfo{person}{Oyvind Tafjord}, \bibinfo{person}{Ashish Sabharwal}, \bibinfo{person}{Peter Clark}, {et~al\mbox{.}}} \bibinfo{year}{2022}\natexlab{}.
\newblock \showarticletitle{Lila: A unified benchmark for mathematical reasoning}.
\newblock \bibinfo{journal}{\emph{arXiv preprint arXiv:2210.17517}} (\bibinfo{year}{2022}).
\newblock


\bibitem[OpenAI(2023)]%
        {openai2023gpt4}
\bibfield{author}{\bibinfo{person}{OpenAI}.} \bibinfo{year}{2023}\natexlab{}.
\newblock \bibinfo{title}{GPT-4 Technical Report}.
\newblock
\newblock
\showeprint[arxiv]{2303.08774}~[cs.CL]


\bibitem[Qiao et~al\mbox{.}(2024)]%
        {qiao2024we}
\bibfield{author}{\bibinfo{person}{Runqi Qiao}, \bibinfo{person}{Qiuna Tan}, \bibinfo{person}{Guanting Dong}, \bibinfo{person}{Minhui Wu}, \bibinfo{person}{Chong Sun}, \bibinfo{person}{Xiaoshuai Song}, \bibinfo{person}{Zhuoma GongQue}, \bibinfo{person}{Shanglin Lei}, \bibinfo{person}{Zhe Wei}, \bibinfo{person}{Miaoxuan Zhang}, {et~al\mbox{.}}} \bibinfo{year}{2024}\natexlab{}.
\newblock \showarticletitle{We-Math: Does Your Large Multimodal Model Achieve Human-like Mathematical Reasoning?}
\newblock \bibinfo{journal}{\emph{arXiv preprint arXiv:2407.01284}} (\bibinfo{year}{2024}).
\newblock


\bibitem[Radford et~al\mbox{.}(2021)]%
        {radford2021learning}
\bibfield{author}{\bibinfo{person}{Alec Radford}, \bibinfo{person}{Jong~Wook Kim}, \bibinfo{person}{Chris Hallacy}, \bibinfo{person}{Aditya Ramesh}, \bibinfo{person}{Gabriel Goh}, \bibinfo{person}{Sandhini Agarwal}, \bibinfo{person}{Girish Sastry}, \bibinfo{person}{Amanda Askell}, \bibinfo{person}{Pamela Mishkin}, \bibinfo{person}{Jack Clark}, {et~al\mbox{.}}} \bibinfo{year}{2021}\natexlab{}.
\newblock \showarticletitle{Learning transferable visual models from natural language supervision}. In \bibinfo{booktitle}{\emph{ICML}}. \bibinfo{pages}{8748--8763}.
\newblock


\bibitem[Team et~al\mbox{.}(2024)]%
        {geminiteam2024geminifamilyhighlycapable}
\bibfield{author}{\bibinfo{person}{Gemini Team}, \bibinfo{person}{Rohan Anil}, \bibinfo{person}{Sebastian Borgeaud}, \bibinfo{person}{Jean-Baptiste Alayrac}, \bibinfo{person}{Jiahui Yu}, \bibinfo{person}{Radu Soricut}, \bibinfo{person}{Johan Schalkwyk}, \bibinfo{person}{Andrew~M. Dai}, \bibinfo{person}{Anja Hauth}, \bibinfo{person}{Katie Millican}, \bibinfo{person}{David Silver}, \bibinfo{person}{Melvin Johnson}, \bibinfo{person}{Ioannis Antonoglou}, \bibinfo{person}{Julian Schrittwieser}, \bibinfo{person}{Amelia Glaese}, \bibinfo{person}{Jilin Chen}, \bibinfo{person}{Emily Pitler}, \bibinfo{person}{Timothy Lillicrap}, \bibinfo{person}{Angeliki Lazaridou}, \bibinfo{person}{Orhan Firat}, \bibinfo{person}{James Molloy}, {and} \bibinfo{person}{Michael Isard}.} \bibinfo{year}{2024}\natexlab{}.
\newblock \bibinfo{title}{Gemini: A Family of Highly Capable Multimodal Models}.
\newblock
\newblock
\showeprint[arxiv]{2312.11805}


\bibitem[Touvron et~al\mbox{.}(2023)]%
        {touvron2023llama}
\bibfield{author}{\bibinfo{person}{Hugo Touvron}, \bibinfo{person}{Thibaut Lavril}, \bibinfo{person}{Gautier Izacard}, \bibinfo{person}{Xavier Martinet}, \bibinfo{person}{Marie-Anne Lachaux}, \bibinfo{person}{Timoth{\'e}e Lacroix}, \bibinfo{person}{Baptiste Rozi{\`e}re}, \bibinfo{person}{Naman Goyal}, \bibinfo{person}{Eric Hambro}, \bibinfo{person}{Faisal Azhar}, {et~al\mbox{.}}} \bibinfo{year}{2023}\natexlab{}.
\newblock \showarticletitle{Llama: Open and efficient foundation language models}.
\newblock \bibinfo{journal}{\emph{arXiv preprint arXiv:2302.13971}} (\bibinfo{year}{2023}).
\newblock


\bibitem[Wang et~al\mbox{.}(2024b)]%
        {wang2024measuring}
\bibfield{author}{\bibinfo{person}{Ke Wang}, \bibinfo{person}{Junting Pan}, \bibinfo{person}{Weikang Shi}, \bibinfo{person}{Zimu Lu}, \bibinfo{person}{Mingjie Zhan}, {and} \bibinfo{person}{Hongsheng Li}.} \bibinfo{year}{2024}\natexlab{b}.
\newblock \showarticletitle{Measuring Multimodal Mathematical Reasoning with MATH-Vision Dataset}.
\newblock \bibinfo{journal}{\emph{arXiv preprint arXiv:2402.14804}} (\bibinfo{year}{2024}).
\newblock


\bibitem[Wang et~al\mbox{.}(2024a)]%
        {Qwen2VL}
\bibfield{author}{\bibinfo{person}{Peng Wang}, \bibinfo{person}{Shuai Bai}, \bibinfo{person}{Sinan Tan}, \bibinfo{person}{Shijie Wang}, \bibinfo{person}{Zhihao Fan}, \bibinfo{person}{Jinze Bai}, \bibinfo{person}{Keqin Chen}, \bibinfo{person}{Xuejing Liu}, \bibinfo{person}{Jialin Wang}, \bibinfo{person}{Wenbin Ge}, \bibinfo{person}{Yang Fan}, \bibinfo{person}{Kai Dang}, \bibinfo{person}{Mengfei Du}, \bibinfo{person}{Xuancheng Ren}, \bibinfo{person}{Rui Men}, \bibinfo{person}{Dayiheng Liu}, \bibinfo{person}{Chang Zhou}, \bibinfo{person}{Jingren Zhou}, {and} \bibinfo{person}{Junyang Lin}.} \bibinfo{year}{2024}\natexlab{a}.
\newblock \showarticletitle{Qwen2-VL: Enhancing Vision-Language Model's Perception of the World at Any Resolution}.
\newblock \bibinfo{journal}{\emph{arXiv preprint arXiv:2409.12191}} (\bibinfo{year}{2024}).
\newblock


\bibitem[Wang et~al\mbox{.}(2023)]%
        {wang2023cogvlm}
\bibfield{author}{\bibinfo{person}{Weihan Wang}, \bibinfo{person}{Qingsong Lv}, \bibinfo{person}{Wenmeng Yu}, \bibinfo{person}{Wenyi Hong}, \bibinfo{person}{Ji Qi}, \bibinfo{person}{Yan Wang}, \bibinfo{person}{Junhui Ji}, \bibinfo{person}{Zhuoyi Yang}, \bibinfo{person}{Lei Zhao}, \bibinfo{person}{Xixuan Song}, \bibinfo{person}{Jiazheng Xu}, \bibinfo{person}{Bin Xu}, \bibinfo{person}{Juanzi Li}, \bibinfo{person}{Yuxiao Dong}, \bibinfo{person}{Ming Ding}, {and} \bibinfo{person}{Jie Tang}.} \bibinfo{year}{2023}\natexlab{}.
\newblock \bibinfo{title}{CogVLM: Visual Expert for Pretrained Language Models}.
\newblock
\newblock
\showeprint[arxiv]{2311.03079}


\bibitem[Wei et~al\mbox{.}(2022)]%
        {wei2022chain}
\bibfield{author}{\bibinfo{person}{Jason Wei}, \bibinfo{person}{Xuezhi Wang}, \bibinfo{person}{Dale Schuurmans}, \bibinfo{person}{Maarten Bosma}, \bibinfo{person}{Fei Xia}, \bibinfo{person}{Ed Chi}, \bibinfo{person}{Quoc~V Le}, \bibinfo{person}{Denny Zhou}, {et~al\mbox{.}}} \bibinfo{year}{2022}\natexlab{}.
\newblock \showarticletitle{Chain-of-thought prompting elicits reasoning in large language models}.
\newblock \bibinfo{journal}{\emph{NeurIPS}}  \bibinfo{volume}{35} (\bibinfo{year}{2022}), \bibinfo{pages}{24824--24837}.
\newblock


\bibitem[Ye et~al\mbox{.}(2024)]%
        {ye2024mplugowlmodularizationempowerslarge}
\bibfield{author}{\bibinfo{person}{Qinghao Ye}, \bibinfo{person}{Haiyang Xu}, \bibinfo{person}{Guohai Xu}, \bibinfo{person}{Jiabo Ye}, \bibinfo{person}{Ming Yan}, \bibinfo{person}{Yiyang Zhou}, \bibinfo{person}{Junyang Wang}, \bibinfo{person}{Anwen Hu}, \bibinfo{person}{Pengcheng Shi}, \bibinfo{person}{Yaya Shi}, \bibinfo{person}{Chenliang Li}, \bibinfo{person}{Yuanhong Xu}, \bibinfo{person}{Hehong Chen}, \bibinfo{person}{Junfeng Tian}, \bibinfo{person}{Qi Qian}, \bibinfo{person}{Ji Zhang}, \bibinfo{person}{Fei Huang}, {and} \bibinfo{person}{Jingren Zhou}.} \bibinfo{year}{2024}\natexlab{}.
\newblock \bibinfo{title}{mPLUG-Owl: Modularization Empowers Large Language Models with Multimodality}.
\newblock
\newblock
\showeprint[arxiv]{2304.14178}


\bibitem[Yue et~al\mbox{.}(2024)]%
        {yue2024mmmu}
\bibfield{author}{\bibinfo{person}{Xiang Yue}, \bibinfo{person}{Yuansheng Ni}, \bibinfo{person}{Kai Zhang}, \bibinfo{person}{Tianyu Zheng}, \bibinfo{person}{Ruoqi Liu}, \bibinfo{person}{Ge Zhang}, \bibinfo{person}{Samuel Stevens}, \bibinfo{person}{Dongfu Jiang}, \bibinfo{person}{Weiming Ren}, \bibinfo{person}{Yuxuan Sun}, {et~al\mbox{.}}} \bibinfo{year}{2024}\natexlab{}.
\newblock \showarticletitle{Mmmu: A massive multi-discipline multimodal understanding and reasoning benchmark for expert agi}. In \bibinfo{booktitle}{\emph{CVPR}}. \bibinfo{pages}{9556--9567}.
\newblock


\bibitem[Zhang et~al\mbox{.}(2024a)]%
        {internlmxcomposer2_5}
\bibfield{author}{\bibinfo{person}{Pan Zhang}, \bibinfo{person}{Xiaoyi Dong}, \bibinfo{person}{Yuhang Zang}, \bibinfo{person}{Yuhang Cao}, \bibinfo{person}{Rui Qian}, \bibinfo{person}{Lin Chen}, \bibinfo{person}{Qipeng Guo}, \bibinfo{person}{Haodong Duan}, \bibinfo{person}{Bin Wang}, \bibinfo{person}{Linke Ouyang}, \bibinfo{person}{Songyang Zhang}, \bibinfo{person}{Wenwei Zhang}, \bibinfo{person}{Yining Li}, \bibinfo{person}{Yang Gao}, \bibinfo{person}{Peng Sun}, \bibinfo{person}{Xinyue Zhang}, \bibinfo{person}{Wei Li}, \bibinfo{person}{Jingwen Li}, \bibinfo{person}{Wenhai Wang}, \bibinfo{person}{Hang Yan}, \bibinfo{person}{Conghui He}, \bibinfo{person}{Xingcheng Zhang}, \bibinfo{person}{Kai Chen}, \bibinfo{person}{Jifeng Dai}, \bibinfo{person}{Yu Qiao}, \bibinfo{person}{Dahua Lin}, {and} \bibinfo{person}{Jiaqi Wang}.} \bibinfo{year}{2024}\natexlab{a}.
\newblock \showarticletitle{InternLM-XComposer-2.5: A Versatile Large Vision Language Model Supporting Long-Contextual Input and Output}.
\newblock \bibinfo{journal}{\emph{arXiv preprint arXiv:2407.03320}} (\bibinfo{year}{2024}).
\newblock


\bibitem[Zhang et~al\mbox{.}(2024b)]%
        {zhang2024llavarenhancedvisualinstruction}
\bibfield{author}{\bibinfo{person}{Yanzhe Zhang}, \bibinfo{person}{Ruiyi Zhang}, \bibinfo{person}{Jiuxiang Gu}, \bibinfo{person}{Yufan Zhou}, \bibinfo{person}{Nedim Lipka}, \bibinfo{person}{Diyi Yang}, {and} \bibinfo{person}{Tong Sun}.} \bibinfo{year}{2024}\natexlab{b}.
\newblock \bibinfo{title}{LLaVAR: Enhanced Visual Instruction Tuning for Text-Rich Image Understanding}.
\newblock
\newblock
\showeprint[arxiv]{2306.17107}


\bibitem[Zhang et~al\mbox{.}(2023)]%
        {zhang2023automatic}
\bibfield{author}{\bibinfo{person}{Zhuosheng Zhang}, \bibinfo{person}{Aston Zhang}, \bibinfo{person}{Mu Li}, {and} \bibinfo{person}{Alex Smola}.} \bibinfo{year}{2023}\natexlab{}.
\newblock \showarticletitle{Automatic Chain of Thought Prompting in Large Language Models}. In \bibinfo{booktitle}{\emph{ICLR}}.
\newblock


\bibitem[Zhu et~al\mbox{.}(2023)]%
        {zhu2023minigpt}
\bibfield{author}{\bibinfo{person}{Deyao Zhu}, \bibinfo{person}{Jun Chen}, \bibinfo{person}{Xiaoqian Shen}, \bibinfo{person}{Xiang Li}, {and} \bibinfo{person}{Mohamed Elhoseiny}.} \bibinfo{year}{2023}\natexlab{}.
\newblock \showarticletitle{Minigpt-4: Enhancing vision-language understanding with advanced large language models}.
\newblock \bibinfo{journal}{\emph{arXiv preprint arXiv:2304.10592}} (\bibinfo{year}{2023}).
\newblock


\end{thebibliography}





\end{document}